\documentclass[sigconf]{acmart}

\AtBeginDocument{%
  }

\usepackage{url}
\usepackage{booktabs}       
\usepackage{nicefrac}       
\usepackage{microtype}      
\usepackage{graphicx}
\usepackage{tabularx} 
\usepackage{enumitem}
\usepackage{wrapfig}
\usepackage{colortbl}
\usepackage{marvosym}
\usepackage{tikz}
\usetikzlibrary{bayesnet}
\PassOptionsToPackage{table}{xcolor}
\usepackage[table]{xcolor}
\usepackage{multirow} 
\usepackage{algorithm}
\usepackage{algorithmic}
\usepackage{tcolorbox}
\usepackage{caption} 
\usepackage{xspace}
\usepackage{setspace}

\usepackage{etoc}
\usepackage{thmtools}
\usepackage{fvextra}

\definecolor{darkblue}{rgb}{0, 0, 0.5}

\hypersetup{
  colorlinks=true,
  citecolor=darkblue,
  linkcolor=darkblue,
  urlcolor=darkblue
}

\setlist[itemize]{leftmargin=*, itemsep=0.5pt, topsep=0.5pt, parsep=0pt, partopsep=0pt}

\definecolor{cvprblue}{rgb}{0.21,0.49,0.74}

\usepackage{etoc}
\usepackage{thmtools}
\definecolor{MyBlue}{HTML}{E6F0FA}
\definecolor{MyBlueA}{HTML}{D6EAF8}     
\definecolor{MyBlueB}{HTML}{CCE5FF}     
\definecolor{MyBlueC}{HTML}{BDD7EE}     
\definecolor{MyPurpleA}{HTML}{F3E6FF} 
\definecolor{MyPurpleB}{HTML}{E6E6FA}    
\definecolor{MyPurpleC}{HTML}{ECE0F9}      

\declaretheoremstyle[
  headfont=\normalfont\bfseries,
  bodyfont=\normalfont,
  spaceabove=6pt, spacebelow=6pt
]{mystyle}

\tcbuselibrary{skins, breakable}
\definecolor{myblue}{RGB}{147,204,255}  
\definecolor{mylightblue}{HTML}{E6F0FA}
\tcbset{
  enhanced, 
  colback=white!200!black, 
  colframe=mylightblue, 
  colbacktitle=mylightblue, 
  title filled, 
  coltitle=myblue, 
  fonttitle=\bfseries, 
  arc=3mm, 
  outer arc=3mm, 
  boxrule=0.5mm, 
  toprule=0.5mm, 
  bottomrule=0.5mm, 
  titlerule=0.5mm, 
  drop fuzzy shadow, 
}

\setlist[itemize]{topsep=1pt, itemsep=0.5pt}

\newtheorem{definition}{Definition}
\newtheorem{problem}{Problem}

\usepackage{bm}         
\newcommand{\model}{\textsc{LlmSynthor}\xspace}

\definecolor{myblue}{HTML}{3A86FF}
\definecolor{mygreen}{HTML}{219653}

\newcommand{\textmyblue}[1]{\textcolor{myblue}{\emph{#1}}}
\newcommand{\textmygreen}[1]{\textcolor{mygreen}{#1}}

\AtEndPreamble{%
  \hypersetup{
    colorlinks=false,
    citecolor=violet,
    linkcolor=violet,
    filecolor=myblue
  }%
}

\copyrightyear{2026}
\acmYear{2026}
\setcopyright{cc}
\setcctype{by-nc-nd}
\acmConference[KDD 2026] {Proceedings of the 32nd ACM SIGKDD Conference on Knowledge Discovery and Data Mining V.2}{August 9--13, 2026}{Jeju Island, Republic of Korea.}
\acmBooktitle{Proceedings of the 32nd ACM SIGKDD Conference on Knowledge Discovery and Data Mining V.2 (KDD 2026), August 9--13, 2026, Jeju Island, Republic of Korea}
\acmISBN{979-8-4007-2259-2/2026/08}
\acmDOI{10.1145/3770855.3818965}
\setcopyright{cc}
\setcctype{by}



\begin{document}

\title{LLMSynthor: Macro-Aligned Micro-Records Synthesis with Large Language Models}

\author{Yihong Tang}
\affiliation{%
  \institution{McGill University}
  \city{Montreal}
  \state{Quebec}
  \country{Canada}
}
\email{yihong.tang@mail.mcgill.ca}

\author{Menglin Kong}
\affiliation{%
  \institution{McGill University}
  \city{Montreal}
  \state{Quebec}
  \country{Canada}
}
\email{menglin.kong@mail.mcgill.ca}

\author{Junlin He}
\affiliation{%
  \institution{The Hong Kong Polytechnic University}
  \city{Hong Kong SAR}
  \country{China}
}
\email{junlin.he@polyu.edu.hk}

\author{Tong Nie}
\affiliation{%
  \institution{The Hong Kong Polytechnic University}
  \city{Hong Kong SAR}
  \country{China}
}
\email{tong.nie@connect.polyu.hk}

\author{Wei Ma}
\affiliation{%
  \institution{The Hong Kong Polytechnic University}
  \city{Hong Kong SAR}
  \country{China}
}
\email{wei.w.ma@polyu.edu.hk}

\author{Lijun Sun}
\authornote{Corresponding author.}
\affiliation{%
  \institution{McGill University}
  \city{Montreal}
  \state{Quebec}
  \country{Canada}
}
\email{lijun.sun@mcgill.ca}

\renewcommand{\shortauthors}{Yihong Tang et al.}

\begin{abstract}
Macro-aligned micro-records are essential for simulations in social science and urban studies. For instance, epidemic models of urban disease spread are only credible when micro-level records reproduce realistic individual mobility and contact patterns, while macro-level aggregates match macro-statistics such as case counts or travel flows. Still, collecting large-scale fine-grained data is often impractical, leaving researchers with only macro-statistics. While Large Language Models (LLMs) can generate realistic micro-records using rich real-world priors learned from vast corpora, naive record-by-record sampling is inefficient and fails to enforce alignment with target macro-statistics. Given this, we propose \textbf{\model}, a framework that transforms a pre-trained LLM into a macro-aware simulator capable of synthesizing realistic micro-records aligned with given macro-statistics. \model incrementally constructs a synthetic dataset by iteratively generating batches of micro-records that reduce discrepancies between synthetic and target macro-statistics. By treating the LLM as a nonparametric copula over joint variable dependencies, the framework ensures alignment with target marginals and joint distributions. To improve efficiency, we introduce LLM Proposal Sampling, where the LLM generates discrepancy-guided proposals specifying variable constraints and record counts. Evaluations on synthetic and real-world datasets (mobility, e-commerce, population) encompassing diverse formats and settings show that \model achieves high record realism, statistical fidelity, and practical utility, positioning it broadly applicable across economics, social science, urban studies, and beyond. Source codes of \model are available at \href{https://github.com/YihongT/LLMSynthor}{https://github.com/YihongT/LLMSynthor}.
\end{abstract}


\maketitle

\section{Introduction}

High-stakes decisions in domains like public health and urban planning are increasingly supported by agent-based simulations of complex human behavior. 
At the micro-level, individual records capture behavioral detail such as mobility and contact patterns, which drive realistic dynamics. At the macro-level, aggregated statistics ensure consistency with population-level trends. 
Only when micro-records are collectively aligned with real-world macro-statistics can such simulations yield valid insights~\cite{jiang2024large}, yet these data are \textmyblue{unattainable} because large-scale collection is infeasible due to both prohibitive costs and stringent privacy constraints~\cite{bellovin2019privacy}. 
Consequently, researchers and policymakers must rely on macro-statistics, such as census reports, leaving a critical gap between macro-level observations and micro-level representations~\cite{zhou2022creating}.
The challenge, therefore, is to synthesize realistic micro-records that are statistically faithful to the available macro-statistics.

\begin{figure}[t]
    \centering
    \includegraphics[width=\linewidth]{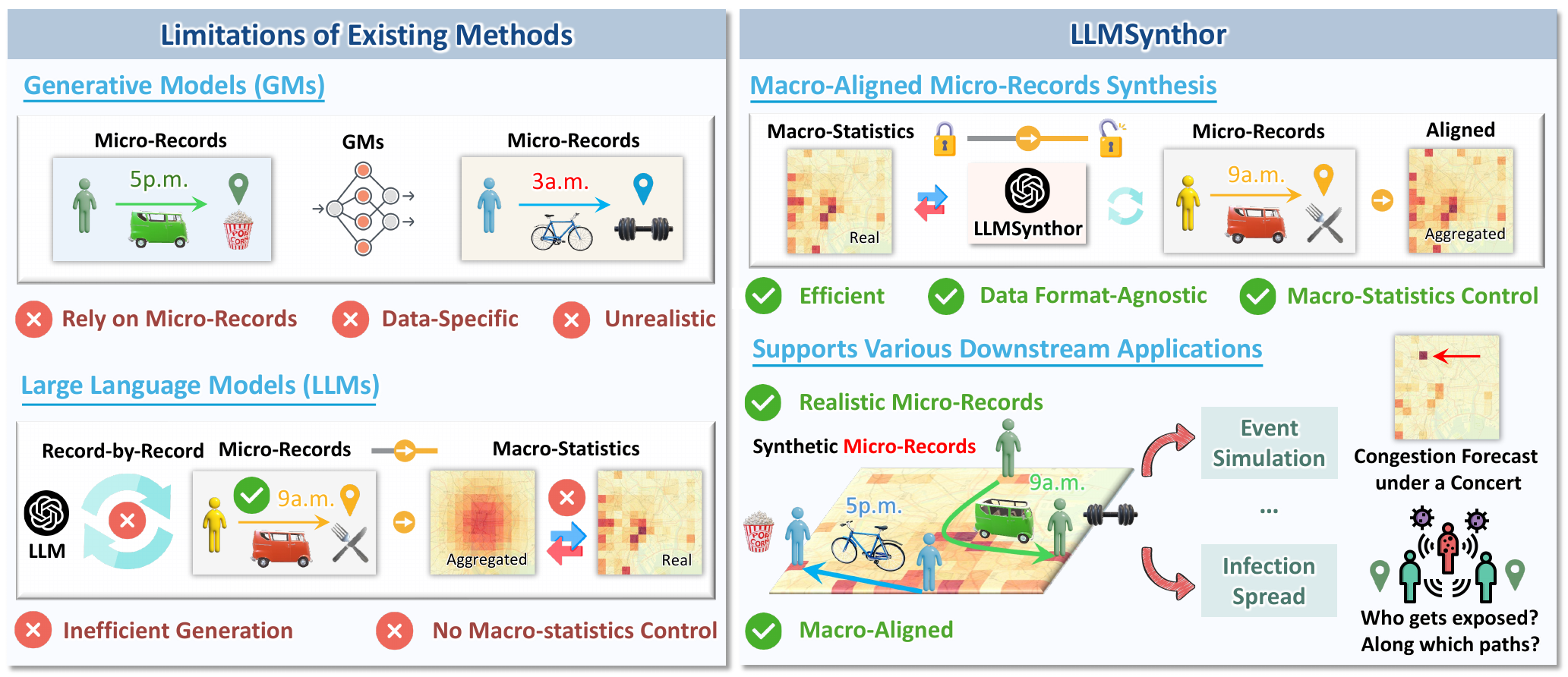}
    \vspace{-10pt}
    \caption{Comparison of generative approaches and \model.
Left: Existing methods either rely on micro-records (Generative Models) or cannot enforce macro-level consistency (LLMs), leading to unrealistic or inefficient generation.
Right: \model synthesizes realistic micro-records directly from macro-statistics, enabling efficient, format-agnostic, and macro-aligned generation.}
    \vspace{-8pt}
    \label{fig:intro}
\end{figure}

This micro-macro synthesis task, however, is beyond the capabilities of existing generative paradigms. Current methods, from classical statistical models to modern deep generative networks~\cite{goodfellow2014generative,ho2020denoising}, are ill-equipped, as they all require access to a large volume of micro-records that are unavailable for model fitting. Furthermore, their reliance on rigid parametric assumptions or implicit model biases often leads to the generation of unrealistic records, such as a six-year-old with a doctorate, necessitating inefficient post-hoc fixes like rejection sampling. Most of these approaches also lack the generality to handle the heterogeneous or unstructured data common in social sciences and urban studies~\cite{ma2020vaem,tang2022domain}, or require extensive manual engineering~\cite{sun2018hierarchical}. This highlights the need for a new framework that can synthesize realistic and complex micro-records that are statistically grounded, guided solely by macro-statistics.

The emergence of Large Language Models (LLMs) offers a promising yet insufficient solution. Harnessing rich real-world priors learned from vast corpora, they excel at generating realistic, complex, and even unstructured micro-records without requiring any fine-tuning, making them powerful universal generative priors~\cite{achiam2023gpt, kojima2022large, kwon2024language}. However, existing approaches fall short in practice. Fine-tuning LLMs requires large volumes of micro-records and substantial computational resources, which are often unavailable in practice. In-context or few-shot prompting can condition local generations, but it still operates record by record, making the process both \textmyblue{inefficient and incapable of enforcing macro-statistical alignment}~\cite{wang2024beyond}. This limitation becomes critical at scale, where population-level simulations demand the rapid generation of large synthetic datasets that are not only realistic at the micro level but also statistically faithful in aggregate. Consequently, standard LLM-based generation may be time-consuming and yield datasets that appear plausible individually yet deviate significantly from target macro-statistics. The fundamental challenge is to enable LLMs to generate large-scale synthetic data that remains both computationally efficient and globally consistent with target macro-statistics.

To address these challenges, we present \model, a framework that transforms a pre-trained LLM into a macro-aware simulator for micro-record synthesis. \model incrementally constructs a synthetic dataset through an iterative feedback loop guided solely by target macro-statistics.
At each iteration, \model quantifies the discrepancy between the synthetic and target macro-statistics, and prompts the LLM to generate a corrective batch of micro-records that reduces this gap. This process is enabled by two core technical innovations. \textmyblue{First}, we interpret the LLM as a powerful nonparametric copula, enabling it to capture complex, nonlinear joint dependencies across variables without imposing rigid statistical assumptions. This allows \model to align the synthetic data with the target by matching all \emph{available} marginal and joint macro-statistics. \textmyblue{Second}, to overcome generation efficiency bottlenecks, we introduce LLM Proposal Sampling, where the LLM creates a generation plan of micro-record proposals, each defining a localized joint distribution over all variables with an associated generation count. This approach uniquely combines the LLM's rich prior knowledge for micro-level realism with rigorous, macro-level statistical control.
Figure~\ref{fig:intro} contrasts our framework with existing paradigms. In summary, our main contributions are:
\begin{itemize}[leftmargin=*]
\item We introduce \model, which transforms a pre-trained LLM into a macro-aware simulator for synthesizing micro-records that preserve realism while aligning with target macro-statistics.
\item We propose two core methodological insights: (i) a nonparametric copula interpretation of LLMs for modeling joint variable dependencies, and (ii) LLM Proposal Sampling for efficient, large-scale, and controlled data generation.
\item We design a discrepancy-guided synthesis process that iteratively extends and aligns the synthetic and the target macro-statistics, providing rigorous, dataset-level statistical control.
\item Evaluations on synthetic and real-world datasets (mobility, e-commerce, population) across diverse formats and settings show that \model consistently outperforms expert baselines in record realism, statistical fidelity, and practical utility, while maintaining broad versatility.
\end{itemize}

\section{Related Work}

\vspace{-1pt}
\paragraph{Dataset Synthesis}

Early data synthesis methods focused on explicit statistical control, using techniques like iterative proportional fitting (IPF)~\cite{beckman1996creating}, Bayesian networks~\cite{sun2018hierarchical}, and copula models~\cite{nelsen2006introduction, okhrin2017copulae} to match marginals and preserve dependencies. While interpretable, these methods often rely on strong assumptions and face challenges with scalability and heterogeneity. Deep generative models, including VAEs~\cite{apellaniz2024improved, tazwar2024tab}, GANs~\cite{xu2019modeling, baowaly2019synthesizing}, and recent diffusion- or flow-based models~\cite{kotelnikov2023tabddpm, kamthe2021copula, zhang2023mixed}, have improved realism and high-dimensional modeling. However, they tend to entangle marginals with dependencies and often require expensive retraining for new domains. LLM-based methods, such as GReaT~\cite{borisov2022language} and HARMONIC~\cite{wang2024harmonic}, treat structured data as natural language, enabling zero-shot transfer and broad domain coverage through autoregressive decoding. However, these methods lack direct control over marginal and joint distributions, are sample inefficient, and struggle to scale with large or heterogeneous datasets.

\vspace{-8pt}
\paragraph{LLMs as Data Generators}
LLMs have shown exceptional versatility as data generators across various domains. They have been used to augment data~\cite{ding2024data}, create instruction-finetuning datasets~\cite{wang2022self, li2024selective}, generate tabular data~\cite{wang2024harmonic, borisov2022language}, synthesize executable code~\cite{nijkamp2022codegen, mankowitz2023faster}, and produce personal mobility data aligned with user preferences~\cite{tang2024itinera} and routines~\cite{jiawei2024large}. LLMs also generate question-answering pairs to enhance model robustness~\cite{chowdhury2023generative,tang2026street} and privacy-preserving text via topic modeling~\cite{tan2025synthesizing}. While existing methods excel at ensuring the semantic or functional quality of individual records, they often fail to control global statistical properties of the dataset. Some works introduce limited control, such as ensuring topic completeness or data coverage~\cite{zhang2023effective}. 
In parallel, recent work studies LLMs as attributed training-data generators and behavioral simulators, focusing on diversity, bias, and persona alignment rather than explicit aggregate control~\cite{yu2023large,holt2025g}.
However, these approaches lack explicit macroscopic statistical control. Most methods generate data record by record or in small, independent batches, without enforcing global statistical properties, highlighting the gap between generating realistic micro-records and ensuring statistical fidelity across the entire dataset, which is directly addressed by \model.

\vspace{-2pt}
\section{Methodology}

\begin{definition}[Micro-record and Dataset]
\label{def:micro_record}
Let $\mathcal{V}$ be a predefined set of variables. A \textbf{micro-record} $x$ is defined as a set of variable-value pairs:
$x=\{(v_i, a_i)\}_{i=1}^{d_x},$
where $v_i \in \mathcal{V}$ and $a_i$ is its value. A \textbf{dataset} is defined as a collection of such records, denoted by $\mathcal{D}=\{x_j\}_{j=1}^{|\mathcal{D}|}$.
\end{definition}

\begin{definition}[Macro-statistics]
\label{def:macro_statistics}
Let $\mathcal{D}$ be a dataset of micro-records. We define an operator \(\Phi\) that maps the dataset into a set of macro-statistics \(\mathcal{S}\), denoted as:
$\Phi: \mathcal{D} \mapsto \mathcal{S}, \text{where } \mathcal{S} = \{\phi_i\}_{i=1}^{|\mathcal{S}|}.$ Here, each element \(\phi_i\) represents a specific aggregated statistic derived from $\mathcal{D}$.
\end{definition}

\begin{figure*}[t]
    \centering
    \includegraphics[width=\linewidth]{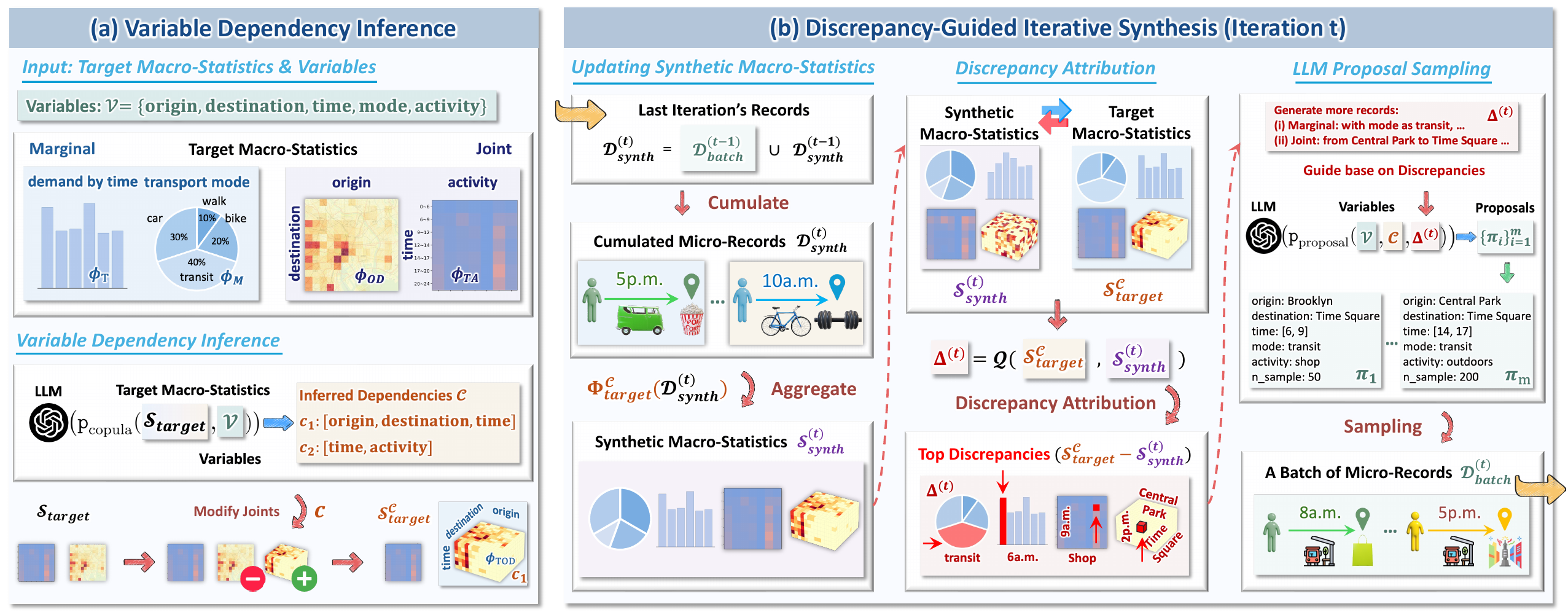}
    \vspace{-20pt}
    \caption{Overview of \model. The system (a) first uses an LLM to infer key variable dependencies, and then modifies the target macro-statistics to align with these inferred dependencies; (b) it then performs discrepancy-guided iterative synthesis. In each iteration, the newly generated micro-records are aggregated to form updated synthetic macro-statistics, which are compared with the target. The differences are attributed to specific variable groups, and guide the LLM to propose new micro-records that correct discrepancies. This feedback loop gradually aligns the synthetic data with the target macro-statistics.}
    \vspace{-10pt} 
    \label{fig:framework}
\end{figure*}
\begin{problem}[Macro-aligned Micro-records Synthesis]
\label{prob:formulation}
Given a set of target macro-statistics $\mathcal{S}_{\mathrm{target}}$, the objective is to construct a dataset of $n$ realistic micro-records, 
$\mathcal{D}_{\mathrm{synth}} = \{\hat{x}_j\}_{j=1}^{n}$, such that the macro-statistics induced from $\mathcal{D}_{\mathrm{synth}}$ 
closely align with $\mathcal{S}_{\mathrm{target}}$. Formally, we seek
$\min_{\mathcal{D}_{\mathrm{synth}}} 
Q\bigl(\Phi(\mathcal{D}_{\mathrm{synth}}), \; \mathcal{S}_{\mathrm{target}}\bigr),$
where $\Phi$ is the aggregation operator defined in Definition~\ref{def:macro_statistics}, 
and $Q$ is a suitable discrepancy measure in the macro-statistics space. 
\end{problem}

In this work, a \emph{micro-record} represents a data record corresponding to an individual entity. 
Each micro-record consists of a set of variables and their associated values. 
To accommodate the flexibility of unstructured data, the number of variables $d_x$ is allowed to vary across records. \emph{Macro-statistics} summarize the distributional properties of a dataset and serve as the alignment targets for synthetic micro-records. In practice, macro-statistics are often the only information available from external sources. Formally, the statistics in $\mathcal{S}$ can be categorized into two types: \textmyblue{\emph{(1) Marginal statistics}}, which describe the distribution of a single variable (e.g., a frequency vector of education levels), and
\textmyblue{\emph{(2) Joint statistics}}, which capture dependencies among multiple variables (e.g., a contingency table of education by employment status). While the aggregation operator $\Phi$ can, in principle, compute arbitrary statistics, its specific instantiation is application-dependent and determined by the macro-statistics available in a given context.

Consider a mobility dataset. A micro-record may represent a single trip, for example \textmygreen{$x = \{(\texttt{origin}$, $\text{`Times Square'})$, $(\texttt{destination}$, $\text{`Central Park'})$, $(\texttt{mode}, \text{`Bike'})$, $(\texttt{time}, 17)\}$}. Here, the variable set is \textmygreen{$\mathcal{V} = \{\texttt{origin}, \texttt{destination}, \texttt{mode}, \texttt{time}\}$}, and each micro-record assigns concrete values to these variables. Alternatively, a micro-record may be \textmyblue{unstructured}, such as an entire daily travel diary, where the number of trips varies across individuals.
In real-world settings, only macro-statistics are available from external sources, denoted by $\mathcal{S}_{\mathrm{target}}^{\mathrm{mob}}$. These may include \textmyblue{marginal statistics}, such as the frequency distribution of transportation modes, and \textmyblue{joint statistics}, such as a contingency table over origin-destination pairs. Accordingly, the aggregation operator $\Phi^{\mathrm{mob}}_{\mathrm{target}}$, defined in Definition~\ref{def:macro_statistics}, is instantiated for the mobility domain so that $\Phi^{\mathrm{mob}}_{\mathrm{target}}(\mathcal{D}^{\mathrm{mob}}_{\mathrm{synth}})$ produces statistics directly comparable to $\mathcal{S}_{\mathrm{target}}^{\mathrm{mob}}$. The objective is therefore to construct a synthetic dataset of micro-records $\mathcal{D}^{\mathrm{mob}}_{\mathrm{synth}}$ such that $\Phi^{\mathrm{mob}}_{\mathrm{target}}(\mathcal{D}^{\mathrm{mob}}_{\mathrm{synth}})$ closely matches the target macro-statistics $\mathcal{S}_{\mathrm{target}}^{\mathrm{mob}}$. For instance, if the target specifies that 30\% of trips are made by bike (a marginal statistic) and 10\% travel from Times Square to Central Park (a joint statistic), then the aggregated synthetic dataset must reproduce these proportions.

\subsection{Overview}

\model comprises two core components:
(i) \emph{Variable Dependency Inference}, in which the LLM serves as a copula-like mechanism to capture dependencies among variables and identify informative joint macro-statistics beyond those explicitly provided; and
(ii) \emph{Discrepancy-Guided Iterative Synthesis}, where, at each iteration \(t\), the LLM generates a batch of micro-records to progressively reduce the discrepancy \(Q\).
An overview of the framework is shown in Figure~\ref{fig:framework}, while a simplified illustrative example of the synthesis process is provided in Figure~\ref{fig:example_full} in Appendix~\ref{appx:examples}.

\subsection{Variable Dependency Inference}

\label{sec:copula}

\model operates exclusively on macro-statistics, which provide complementary but incomplete views of the underlying data distribution. While marginal statistics are indispensable for preserving the basic distributions of individual variables, joint macro-statistics may be either redundant or missing, depending on data availability and collection constraints. The goal of dependency inference is to identify an \emph{informative set of variable combinations} whose dependencies are essential for faithful synthesis.

We draw inspiration from copula theory~\cite{sklar1959fonctions}, which decomposes a multivariate distribution into marginal distributions and a dependency structure. 
Departing from parametric copula models, we interpret a pre-trained LLM as a \emph{nonparametric, semantics-aware copula} that can infer salient dependency structures directly from macro-level information. Specifically, given the variable set $\mathcal{V}$, the LLM infers a collection of variable subsets $\mathcal{C} = \{c_k\}_{k=1}^{|\mathcal{C}|}$, where each $c_k \subseteq \mathcal{V}$ represents a group of variables expected to exhibit strong statistical dependence.
This inference leverages both (i) the semantic relationships encoded in variable names (e.g., ``education'' and ``income'') and (ii) the available macro-statistics $\mathcal{S}_{\mathrm{target}}$:
\vspace{-1pt}
\begin{equation}\label{eq:copulallm}
\mathcal{C} \sim \texttt{LLM}\big(\texttt{p}_{\mathrm{copula}}(\mathcal{S}_{\mathrm{target}}, \mathcal{V})\big),
\end{equation}
\vspace{-12pt}

\noindent where the prompt $\texttt{p}_{\mathrm{copula}}$ elicits informative variable combinations (see Appendix~\ref{sec:prompt_copula}).
For example, given the variable set \textmygreen{$\mathcal{V} = \{\texttt{origin}$, $ \texttt{destination}, \texttt{mode}, \texttt{time}\}$}, the LLM may infer dependency subsets \textmygreen{$\{c_1 = [\texttt{origin}, \texttt{destination}, \texttt{time}], \; c_2 = [\texttt{time}, \texttt{activity}]\}$}, reflecting time-dependent travel demand and activity patterns.

Having inferred $\mathcal{C}$, we retain all marginal statistics and selectively expand or filter joint macro-statistics in $\mathcal{S}_{\mathrm{target}}$ to obtain an informative target set $\mathcal{S}_{\mathrm{target}}^{\mathcal{C}}$. In practice, $\mathcal{S}_{\mathrm{target}}^{\mathcal{C}}$ is restricted to joint statistics that are explicitly available or can be reliably specified by practitioners. When certain inferred dependencies are not supported by available macro-statistics, they may be excluded from $\mathcal{C}$. Alternatively, in settings where approximate control is acceptable, LLMs can be used as an optional mechanism to suggest or approximate missing joint statistics. The aggregation operator is then updated as $\Phi_{\mathrm{target}} \mapsto \Phi_{\mathrm{target}}^{\mathcal{C}}$, ensuring that all induced statistics from any dataset are directly comparable to $\mathcal{S}_{\mathrm{target}}^{\mathcal{C}}$.

Figure~\ref{fig:framework}(a) illustrates this process. In the mobility example, the LLM identifies a dependency $c_1$ over \textmygreen{$[\texttt{origin}, \texttt{destination}, \texttt{time}]$}, leading to the replacement of the original OD statistic $\phi_{\mathrm{OD}}$ with a time-conditioned statistic $\phi_{\mathrm{TOD}}$. By contrast, the dependency \textmygreen{$[\texttt{time}, \texttt{activity}]$} is already supported by available macro-statistics, and the corresponding statistic $\phi_{\mathrm{TA}}$ is preserved.

\subsection{Discrepancy-Guided Iterative Synthesis}
\label{sec:3.3}

With the refined target macro-statistics $\mathcal{S}_{\mathrm{target}}^{\mathcal{C}}$ and the corresponding aggregation operator $\Phi^{\mathcal{C}}_{\mathrm{target}}$ in place, \model enters a discrepancy-guided iterative synthesis loop that progressively constructs the synthetic dataset $\mathcal{D}_{\mathrm{synth}}$. 
At each iteration $t$, \textmyblue{(i)} \model first aggregates the current cumulative dataset to obtain synthetic macro-statistics $\mathcal{S}_{\mathrm{synth}}^{(t)} = \Phi^{\mathcal{C}}_{\mathrm{target}}(\mathcal{D}_{\mathrm{synth}}^{(t)})$, and \textmyblue{(ii)} measures the discrepancy $\Delta^{(t)} = Q(\mathcal{S}_{\mathrm{target}}^{\mathcal{C}}, \mathcal{S}_{\mathrm{synth}}^{(t)})$ between the synthetic and target statistics. 
Conditioned on these discrepancy signals, \textmyblue{(iii)} the LLM generates a \emph{batch} of micro-records $\mathcal{D}_{\mathrm{batch}}^{(t)}$ that is designed to reduce $\Delta^{(t)}$, after which the cumulative dataset is updated as $\mathcal{D}_{\mathrm{synth}}^{(t+1)} = \mathcal{D}_{\mathrm{synth}}^{(t)} \cup \mathcal{D}_{\mathrm{batch}}^{(t)}$. 
This closed-loop process repeats until the synthetic dataset achieves satisfactory alignment with the target macro-statistics.
We now elaborate on each step.

\paragraph{1. Updating Synthetic Macro-statistics.}

At the beginning of iteration $t$, the current synthetic dataset $\mathcal{D}^{(t)}_{\mathrm{synth}}$ is aggregated into macro-statistics using the target-aligned operator $\mathcal{S}^{(t)}_{\mathrm{synth}} = \Phi^{\mathcal{C}}_{\mathrm{target}}\!\left(\mathcal{D}^{(t)}_{\mathrm{synth}}\right)$. This step produces a macro-level summary of the cumulative synthetic data that is directly comparable to the target statistics.

To ensure consistency across heterogeneous variables, each micro-record $x \in \mathcal{D}^{(t)}_{\mathrm{synth}}$ is mapped into a unified discretized space. Discrete variables are represented as frequency vectors, while continuous variables are discretized into bins (e.g., quantile-based) following the same scheme used by the target macro-statistics. For joint statistics involving multiple variables, the corresponding bins define contingency tables. This unified representation ensures that synthetic macro-statistics are \emph{directly comparable} to the target macro-statistics, both in discretization and semantic meaning.

\paragraph{2. Discrepancy Attribution.}

Given the updated synthetic macro-statistics $\mathcal{S}^{(t)}_{\mathrm{synth}}$ and the target macro-statistics $\mathcal{S}_{\mathrm{target}}^{\mathcal{C}}$, we next quantify their discrepancy $Q\bigl(\mathcal{S}_{\mathrm{target}}^{\mathcal{C}},\, \mathcal{S}^{(t)}_{\mathrm{synth}} \bigr)$, defined by:
\begin{equation}\label{eq:discrepancy}
\Delta^{(t)} = \left\{ \delta_1^{(t)}, \delta_2^{(t)}, \dots, \delta_{|\mathcal{S}_{\mathrm{target}}^{\mathcal{C}}|}^{(t)} \right\}
= Q\bigl(\mathcal{S}_{\mathrm{target}}^{\mathcal{C}},\, \mathcal{S}^{(t)}_{\mathrm{synth}} \bigr),
\end{equation}
where $Q(\cdot,\cdot)$ is a discrepancy function applied element-wise to each macro-statistic.
In this work, we implement $Q$ as the directed frequency difference
$\mathcal{S}_{\mathrm{target}}^{\mathcal{C}} - \mathcal{S}^{(t)}_{\mathrm{synth}}$.
Positive values indicate underrepresented portions of the synthetic data relative to the target, while negative values indicate overrepresentation.
These discrepancy signals explicitly identify which macro-level patterns require correction and in which direction. Because $\Phi^{\mathcal{C}}_{\mathrm{target}}(\cdot)$ operates on a discretized representation, each discrepancy is associated with specific bins or variable combinations (e.g., \textmygreen{``time = 5-6 p.m. and mode = bike''}).
As a result, discrepancies are not only quantitative but also \textmyblue{attributable} to concrete subsets of variables, enabling them to be translated into actionable guidance for later generations.

For example, consider the directed discrepancy for transportation modes:
\textmygreen{$\delta^{(t)}_{M} = \{\text{bike}: +30\%, \text{car}: -20\%\}$}.
This indicates that bike trips are underrepresented in the synthetic data by 30\%, motivating the generation of additional bike-related records.
Figure~\ref{666} provides an illustrative example.
Further details are in Appendix~\ref{appx:exp_ipl}.

\paragraph{3. Discrepancy-Guided LLM Proposal Sampling.}

The discrepancy signals $\Delta^{(t)}$ are then used to guide the generation of a new batch of micro-records. Rather than generating records one by one, which is inefficient and difficult to enforce distribution control at the macro-level, we shift the role of the LLM from a direct generator to a high-level \emph{planner}.
Specifically, we introduce \emph{LLM Proposal Sampling}, in which the LLM generates a set of $m$ proposals:
\begin{equation}\label{eq:llmproposal}
\{\pi_{1}, \dots, \pi_{m}\} \sim \texttt{LLM}\bigl(\texttt{p}_{\mathrm{proposal}}(\mathcal{V},\, \mathcal{C},\, \Delta^{(t)})\bigr),
\end{equation}
using a prompt $\texttt{p}_{\mathrm{proposal}}$ that instructs the LLM to design proposals aimed at reducing the identified discrepancies while respecting the inferred dependency structure $\mathcal{C}$.
The complete implemented prompt $\texttt{p}_{\mathrm{proposal}}$ is provided in Appendix~\ref{appx:prompts}.

Each generated proposal $\pi_i$ specifies a localized joint configuration over all variables together with the number of micro-records to generate. For discrete variables, this corresponds to assigning specific categorical values (e.g., \textmygreen{\texttt{mode} = bike}); for continuous variables, it specifies value ranges (e.g., \textmygreen{\texttt{time} $\in$ [17, 20]}), from which values are sampled uniformly for simplicity. By explicitly planning record counts, proposal sampling enables quantitatively targeted generation rather than uncontrolled sampling.

Importantly, all proposals are aligned with the discretization scheme used by $\Phi^{\mathcal{C}}_{\mathrm{target}}$, ensuring that their effects on macro-statistics are predictable. At the same time, the LLM’s learned priors ensure that the proposed joint configurations remain realistic. An example is shown in Figure~\ref{fig:framework}(b), where one proposal \textmygreen{$\pi_1$} specifies 50 records with origin ``Brooklyn'', destination ``Times Square'', mode ``transit'', activity ``Shop'', and time range [6, 9].

Finally, records sampled from these proposals form a batch $\mathcal{D}^{(t)}_{\mathrm{batch}}$, which is merged into the cumulative dataset as $\mathcal{D}^{(t+1)}_{\mathrm{synth}} = \mathcal{D}^{(t)}_{\mathrm{synth}} \cup \mathcal{D}^{(t)}_{\mathrm{batch}}$. By operating at the level of proposals rather than individual records, \model acts as a high-level distributional controller, efficiently translating macro-level discrepancy signals into targeted micro-level generation while preserving realism.

\vspace{-4pt}

\paragraph{Iterative Synthesis.}

\label{sec:loop}
\vspace{-5pt}
\begin{algorithm}[H]
\caption{Iterative Synthesis}
\label{alg:llm_loop}
{\setstretch{.1}
\footnotesize
\begin{algorithmic}[1]
  \REQUIRE Target macro-statistics $\mathcal{S}_{\mathrm{target}}$, aggregation operator $\Phi_{\mathrm{target}}$, discrepancy measure $Q$, iterations $T$
  \STATE $\mathcal{D}^{(0)}_{\mathrm{synth}} \gets \emptyset$ \hfill $\triangleright$ initialize synthetic dataset
  \STATE $\mathcal{C} \gets \texttt{LLM}\bigl(\texttt{p}_{\mathrm{copula}}(\mathcal{S}_{\mathrm{target}}, \mathcal{V})\bigr)$ \hfill $\triangleright$ variable dependency inference
  \STATE $\mathcal{S}_{\mathrm{target}}^{\mathcal{C}} \overset{\mathcal{C}}{\gets} \mathcal{S}_{\mathrm{target}}$; 
         $\Phi_{\mathrm{target}}^{\mathcal{C}} \overset{\mathcal{C}}{\gets} \Phi_{\mathrm{target}}$ \hfill $\triangleright$ refine targets and operator
  \FOR{$t=1$ \TO $T$}
    \STATE $\mathcal{S}^{(t)}_{\mathrm{synth}} \gets \Phi^{\mathcal{C}}_{\mathrm{target}}(\mathcal{D}^{(t)}_{\mathrm{synth}})$ \hfill $\triangleright$ updating synthetic macro-statistics
    \STATE $\Delta^{(t)} \gets Q\bigl(\mathcal{S}_{\mathrm{target}}^{\mathcal{C}},\,\mathcal{S}^{(t)}_{\mathrm{synth}} \bigr)$ \hfill $\triangleright$ discrepancy attribution
    \STATE $\{\pi^{(t)}_i\} \gets \texttt{LLM}\bigl(\texttt{p}_{\mathrm{proposal}}(\mathcal{V},\mathcal{C}, \Delta^{(t)})\bigr)$ \hfill $\triangleright$ plan proposals
    \STATE $\mathcal{D}^{(t)}_{\mathrm{batch}} \gets \bigcup_i \{\hat{x}^{(t)}_i\},\; \hat{x}^{(t)}_i \sim \pi^{(t)}_i$ \hfill $\triangleright$ LLM proposal sampling
    \STATE $\mathcal{D}^{(t+1)}_{\mathrm{synth}} \gets \mathcal{D}^{(t)}_{\mathrm{synth}} \cup \mathcal{D}^{(t)}_{\mathrm{batch}}$ \hfill $\triangleright$ update dataset
  \ENDFOR
  \RETURN $\mathcal{D}^{(T)}_{\mathrm{synth}}$
\end{algorithmic}
}
\end{algorithm}
\vspace{-5pt}

Algorithm~\ref{alg:llm_loop} summarizes the discrepancy-guided iterative synthesis process.
After initializing an empty synthetic dataset and inferring the variable dependency structure $\mathcal{C}$, the algorithm iteratively expands $\mathcal{D}_{\mathrm{synth}}$ through a closed-loop procedure.
\textmyblue{First}, at iteration $t$, the current cumulative synthetic dataset $\mathcal{D}^{(t)}_{\mathrm{synth}}$ is aggregated into macro-statistics using the refined operator $\Phi^{\mathcal{C}}_{\mathrm{target}}$, and compared with the target macro-statistics $\mathcal{S}_{\mathrm{target}}^{\mathcal{C}}$ via the discrepancy measure $Q$, yielding discrepancy signals $\Delta^{(t)}$ that identify which macro-level patterns are under- or over-represented.
\textmyblue{Second}, conditioned on the discrepancy signals $\Delta^{(t)}$ and the inferred dependency structure $\mathcal{C}$, the LLM performs proposal planning by generating a set of proposal distributions $\{\pi^{(t)}_i\}$.
Each proposal specifies a localized joint configuration over variables and a corresponding record count, and is designed to directly target the dominant mismatches revealed by $\Delta^{(t)}$.
\textmyblue{Third}, synthetic records are sampled from these proposals to form a batch $\mathcal{D}^{(t)}_{\mathrm{batch}}$, which is merged into the cumulative dataset to obtain $\mathcal{D}^{(t+1)}_{\mathrm{synth}}$.
By iteratively aggregating, attributing discrepancies, and generating targeted batches, the algorithm progressively reduces macro-level mismatches while preserving micro-level realism, steering the synthetic dataset toward alignment with the target macro-statistics.

\section{Experiments}

To comprehensively evaluate \model, our experiments are structured to answer three research questions (RQs). \textmyblue{RQ1}: How effectively can \model synthesize realistic and usable micro-records from limited aggregate-level macro-statistics? \textmyblue{RQ2}: How does \model's statistical fidelity compare to state-of-the-art models trained on full micro-record datasets? \textmyblue{RQ3}: How effectively can \model synthesize \emph{realistic} and \emph{unstructured} micro-records, without task-specific manual engineering?
We address these RQs using three practical tasks detailed below. We perform experiments using the Chat Completion mode of GPT-4.1-nano~\cite{gpt4-1}.

\subsection{Mobility Synthesis}

Mobility synthesis aims to generate a complete dataset of realistic micro-records, each detailing an individual's time-stamped origin-destination trip, activity, and transport mode. This is an essential capability for urban applications like transport planning and event simulation, as comprehensive, individual-level mobility data for an entire population is impractical to collect.

\noindent \textbf{Task Setup.} To answer \textmyblue{RQ1}, we design a mobility synthesis task that mirrors a common real-world constraint: fusing aggregate information from \textmyblue{multiple} complementary data sources. From Open PFLOW~\cite{kashiyama2017open}, we extract trips (origin, destination, timestamp) and assign transport modes. Since OpenPFLOW lacks activity labels, we incorporate time-activity patterns from LLMob~\cite{jiawei2024large}. This task tests the ability to align spatiotemporal and behavioral data by generating 30,000 trips in a day in Tokyo to match both macro-statistics. As existing methods cannot handle such \textmyblue{Mixed-Source} synthesis without manual adaptations, we focus on a qualitative assessment of \model's unique capabilities. 

\begin{figure}[h]  
  \centering
  \includegraphics[width=\columnwidth]{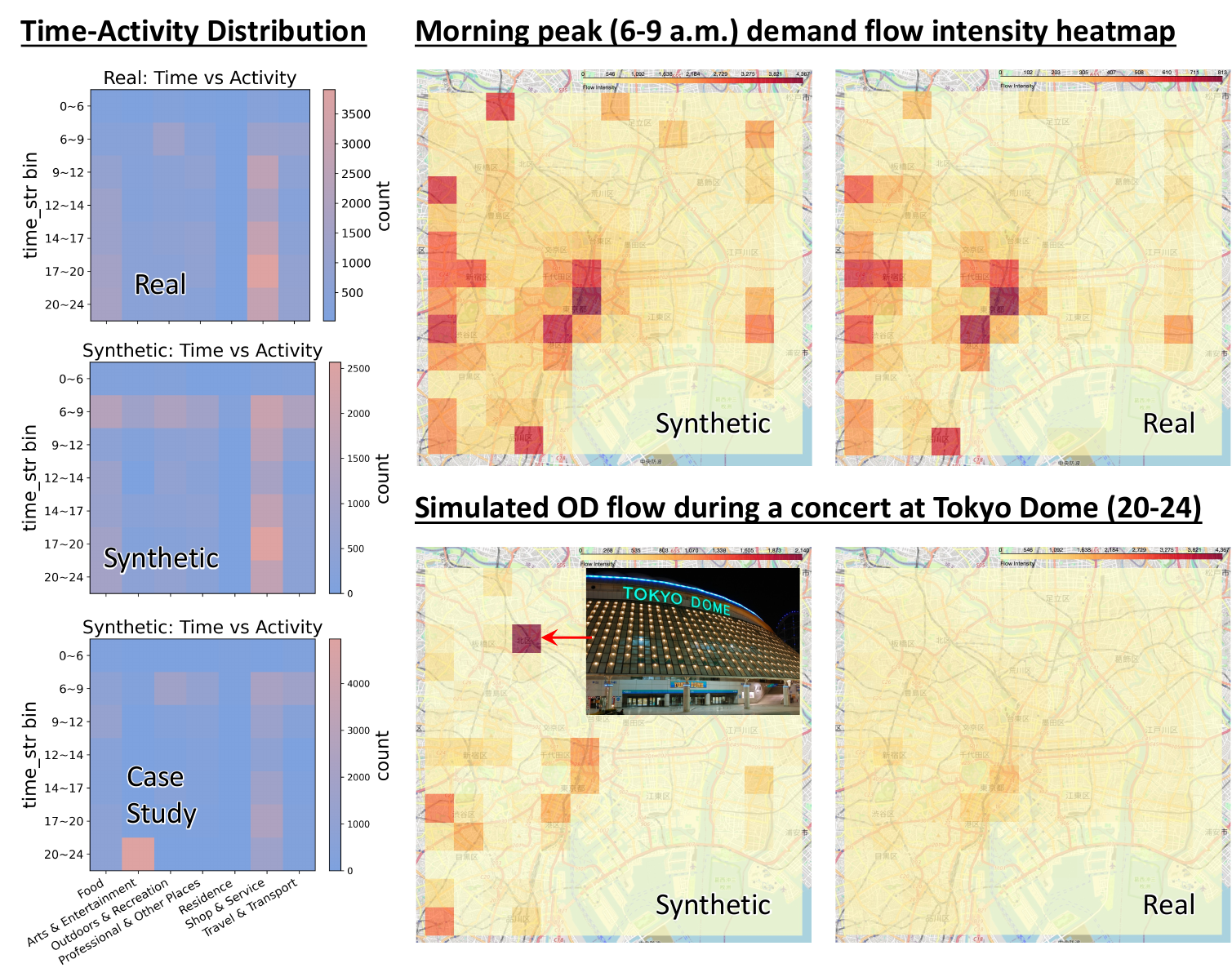}
  \vspace{-24pt}
  \caption{Real vs. synthetic mobility patterns.}
  \vspace{-14pt}
  \label{fig:mob_main}
\end{figure}

\begin{figure}[h]
  \centering
  \vspace{-8pt}
  \includegraphics[width=1\linewidth]{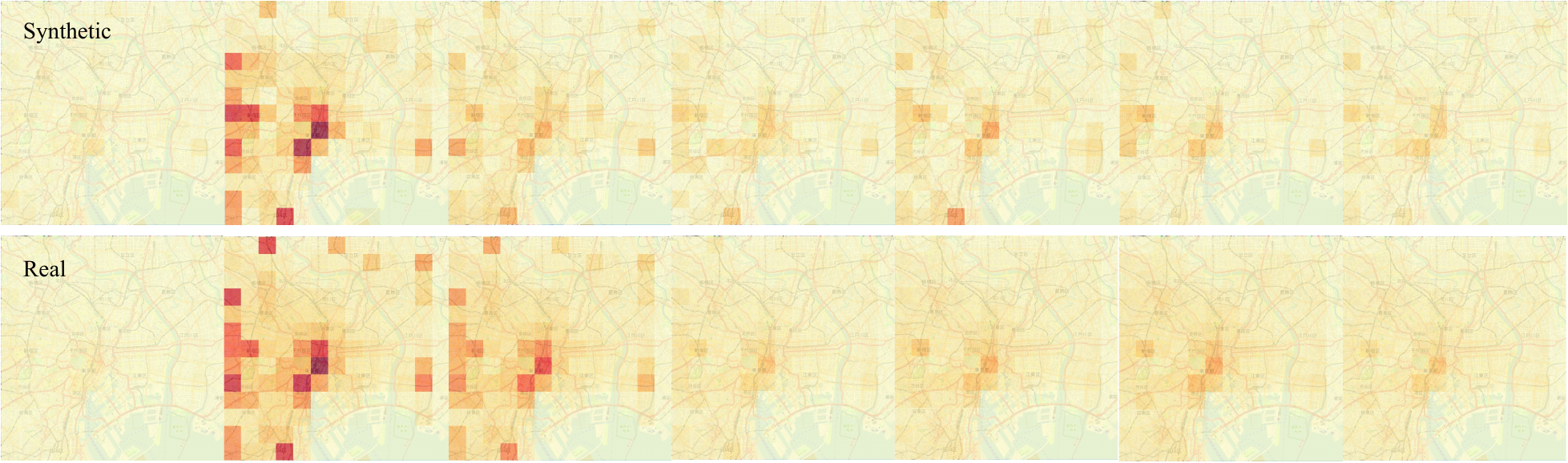}
  \vspace{-22pt}
  \caption{Flow intensity maps for mobility data across seven time intervals: \texttt{0-6}, \texttt{6-9}, \texttt{9-12}, \texttt{12-14}, \texttt{14-17}, \texttt{17-20}, and \texttt{20-24}.}
  \label{fig:mob_st}
  \vspace{-10pt}
\end{figure}

\begin{figure}[h]
  \centering
  \vspace{-10pt}
  \includegraphics[width=.9\linewidth]{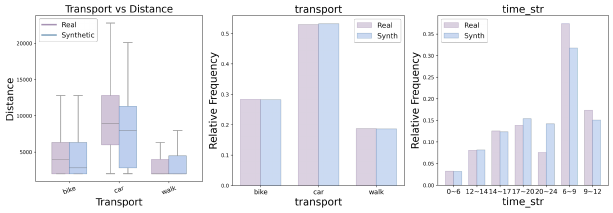}
  \vspace{-13pt}
  \caption{Distribution comparisons for mobility variables.}
  \label{fig:mob_joint}
  \vspace{-5pt}
\end{figure}

\noindent \textbf{Results.} The results provide strong evidence for the first component of \textmyblue{RQ1}: \model's ability to generate realistic micro-records. Figure~\ref{fig:mob_main} compares the synthetic data against the target macro-statistics. The time-activity heatmaps on the left show close alignment, accurately capturing commuting peaks and midday activity rises. The OD flow heatmaps during the morning peak confirm that the synthetic trips reproduce key spatial patterns, matching high-density areas. These findings demonstrate that the synthesized population, in aggregate, successfully reproduces the guiding macro-level patterns. Furthermore, as shown in 
Figures~\ref{fig:mob_st} and~\ref{fig:mob_joint}, the generated micro-records also exhibit realistic internal structures, such as realistic correlations between travel mode and distance, reflecting their micro-level realism.

\noindent \textbf{Controllable Mobility Synthesis for Events Simulation.} To address the second component of \textmyblue{RQ1} concerning the utility of the generated data, we demonstrate a key advantage of our framework: the ability to effortlessly incorporate arbitrary context into the synthesis process. We test \model's controllability in a ``what-if'' scenario. We simulate a concert at Tokyo Dome (20-24h) by simply adding the prompt \textmygreen{\texttt{<} There will be a concert from 20-24 at Tokyo Dome \texttt{>}} during proposal generation. As shown in Figure~\ref{fig:mob_main}, this simple intervention causes \model to generate a surge of trips to the event location while preserving realistic background flows. This demonstrates \model’s potential as a powerful tool for scenario planning, allowing policymakers to simulate the effects of large events using detailed synthetic micro-records.

Figure~\ref{fig:mob_st} shows a detailed comparison of spatial-temporal flow intensity between real and synthetic data across seven time intervals throughout the day. Each map captures the aggregate origin and destination activity within the Tokyo metropolitan area during a specific time window. The synthetic data successfully preserves major spatial patterns such as morning and evening commute flows, also capturing temporal variations in trip density. This highlights the model’s ability to maintain realistic spatiotemporal dynamics.

Figure~\ref{fig:mob_joint} presents additional joint distribution visualizations across key mobility attributes. The left plot illustrates the correlation between transport mode and travel distance, showing that synthetic records preserve realistic distance-dependent mode preferences (e.g., longer trips by car). The middle and right plots show marginal distributions for transport modes and time intervals, further confirming strong alignment between real and synthetic mobility behavior. Together, these results demonstrate that \model can faithfully reproduce both spatial structure and behavioral signals critical for urban simulation and mobility planning.

\vspace{-5pt}

\subsection{E-Commerce Transaction Synthesis}

\renewcommand{\arraystretch}{.96}
\begin{table*}[t]
\resizebox{\linewidth}{!}{
\begin{tabular}{l|c|cc|cc|cc|cc|cc|cc}
\toprule
\multirow{2}{*}{\textbf{Methods}} & \multirow{2}{*}{\textbf{Rej.\% $\downarrow$}} & \textbf{Was $\downarrow$} & \textbf{Gap $\downarrow$} & \textbf{Tvd $\downarrow$} & \textbf{Gap $\downarrow$} & \textbf{Tvd $\downarrow$} & \textbf{Gap $\downarrow$} & \textbf{Tvd $\downarrow$} & \textbf{Gap $\downarrow$} & \textbf{Was $\downarrow$} & \textbf{Gap $\downarrow$} & \textbf{Tvd $\downarrow$} & \textbf{Gap $\downarrow$} \\ \cmidrule(lr){3-4} \cmidrule(lr){5-6} \cmidrule(lr){7-8} \cmidrule(lr){9-10} \cmidrule(lr){11-12} \cmidrule(lr){13-14} 
 &  & \multicolumn{2}{c|}{\textbf{age}} & \multicolumn{2}{c|}{\textbf{gender}} & \multicolumn{2}{c|}{\textbf{location}} & \multicolumn{2}{c|}{\textbf{category}} & \multicolumn{2}{c|}{\textbf{price}} & \multicolumn{2}{c}{\textbf{payment}}  \\ \midrule
\cellcolor{MyPurpleA!15} TVAE~\cite{xu2019modeling} & 1.7 ± 0.2 & 2.06 & 0.032 & \underline{0.008} & 0.01 & 0.056 & 0.043 & 0.054 & 0.02 & 113.194 & 0.085 & 0.017 & 0.013 \\
\cellcolor{MyPurpleA!15} CTGAN~\cite{xu2019modeling} & 5.1 ± 1.1 & 4.429 & 0.057 & 0.117 & 0.065 & 0.162 & 0.076 & 0.080 & 0.028 & 138.998 & 0.059 & 0.088 & 0.022 \\
\cellcolor{MyPurpleA!15} CopulaGAN~\cite{d2017conscientious} & 4.8 ± 1.0 & 4.82 & 0.027 & 0.052 & 0.016 & 0.045 & 0.031 & 0.057 & 0.024 & 151.239 & 0.047 & 0.045 & 0.014 \\
\cellcolor{MyPurpleA!15} GReaT~\cite{borisov2022language} & 0.8 ± 0.1 & 2.862 & 0.052 & 0.016 & \underline{0.009} & 0.039 & 0.020 & 0.045 & 0.027 & 169.866 & 0.104 & 0.009 & 0.012 \\
\cellcolor{MyPurpleA!15} TabSyn~\cite{zhang2023mixed} & 1.9 ± 0.3 & \underline{1.196} & \textbf{0.012} & 0.012 & 0.022 & 0.007 & 0.022 & 0.045 & \textbf{0.01} & 114.12 & 0.067 & 0.028 & \underline{0.005} \\
\cellcolor{MyPurpleA!50} Spada~\cite{yang2025doubling} & 4.1 ± 0.9 & 1.42 & 0.025 & 0.013 & 0.019 & \underline{0.003} & 0.022 & 0.016 & 0.024 & \underline{54.603} & \underline{0.023} & 0.025 & 0.009 \\
\cellcolor{MyPurpleA!100} IPF~\cite{kolenikov2014calibrating} & 3.9 ± 0.6 & 5.27 & 0.087 & 0.018 & \textbf{0.008} & 0.026 & 0.016 & \underline{0.012} & 0.024 & 139.957 & 0.078 & 0.035 & 0.013 \\
\cellcolor{MyPurpleA!100} LLM-ICL~\cite{brown2020language} & \textbf{0.1 ± 0.0} & 20.61 & 0.194 & 0.026 & 0.026 & 0.024 & \underline{0.014} & 0.441 & 0.215 & 279.743 & 0.205 & 0.163 & 0.081 \\ \midrule
\rowcolor{MyBlueA!100} \cellcolor{MyPurpleA!100} \model (ours)  & \underline{0.3 ± 0.1} & \textbf{1.13} & \underline{0.023} & \textbf{0.002} & \textbf{0.008} & \textbf{0.002} & \textbf{0.012} & \textbf{0.010} & 0.022 & \textbf{12.762} & \textbf{0.011} & \textbf{0.003} & \textbf{0.004} \\ \bottomrule
\end{tabular}
}
\vspace{3pt}
\caption{Marginal alignment evaluation ($\downarrow$ is better). 
Methods differ by supervision: IPF, LLM-ICL, and Ours (\model) are purely \textit{macro-supervised}; Spada is \textit{macro+micro supervised}; while others are standard \textit{micro-supervised} models. 
Metrics include Wasserstein distance (Was), Total Variation Distance (Tvd), and the classifier two-sample test gap (Gap, defined as $|\text{Acc} - 0.5|$).}

\label{tab:marginal_trans}
\vspace{-7pt}
\end{table*}

\begin{figure*}[htbp]
  \centering

    \begin{minipage}[h]{0.35\textwidth}
    \renewcommand{\arraystretch}{1.8}
    \centering
    \resizebox{\linewidth}{!}{
    \begin{tabular}{l|cccccc}
    \toprule
    \multirow{2}{*}{\textbf{Methods}} & \textbf{Jsd $\downarrow$} & \textbf{Gap $\downarrow$} & \textbf{Jsd $\downarrow$} & \textbf{Gap $\downarrow$} & \textbf{Jsd $\downarrow$} & \textbf{Gap $\downarrow$} \\ \cmidrule(lr){2-3} \cmidrule(lr){4-5} \cmidrule(lr){6-7}
     &  \multicolumn{2}{c}{$[v_A, v_G, v_C]$} & \multicolumn{2}{c}{$[v_C, v_X]$} & \multicolumn{2}{c}{$[v_L, v_M]$} \\ \midrule
    \cellcolor{MyPurpleA!15} TVAE & 0.23 & 0.074 & 0.245 & 0.106 & 0.185 & 0.051 \\
    \cellcolor{MyPurpleA!15} CTGAN & 0.133 & 0.055 & 0.298 & 0.098 & 0.145 & 0.076 \\
    \cellcolor{MyPurpleA!15} CopulaGAN & 0.133 & 0.057 & 0.280 & 0.102 & 0.069 & 0.018 \\
    \cellcolor{MyPurpleA!15} GReaT & 0.087 & 0.058 & 0.382 & 0.177 & 0.038 & 0.020 \\
    \cellcolor{MyPurpleA!15} TabSyn & \underline{0.083} & \textbf{0.022} & 0.237 & 0.082 & \underline{0.027} & \underline{0.015} \\
    \cellcolor{MyPurpleA!50} Spada & 0.085 & 0.024 & \underline{0.225} & \underline{0.033} & 0.103 & 0.03 \\
    \cellcolor{MyPurpleA!100} IPF & 0.117 & 0.049 & 0.504 & 0.154 & 0.113 & 0.025 \\
    \cellcolor{MyPurpleA!100} LLM-ICL & 0.437 & 0.216 & 0.591 & 0.242 & 0.476 & 0.081 \\ \midrule
    \rowcolor{MyBlueA!100} \cellcolor{MyPurpleA!100} Ours & \textbf{0.071} & \textbf{0.022} & \textbf{0.134} & \textbf{0.020} & \textbf{0.007} & \textbf{0.007} \\ \bottomrule
    \end{tabular}
}
\vspace{10pt}
\captionof{table}{Joint evaluations.}
  \label{tab:joint_trans}
  \end{minipage}
  \hfill
  \begin{minipage}[h]{0.64\textwidth}
    \centering
    \vspace{-16pt}  
    \includegraphics[width=\linewidth]{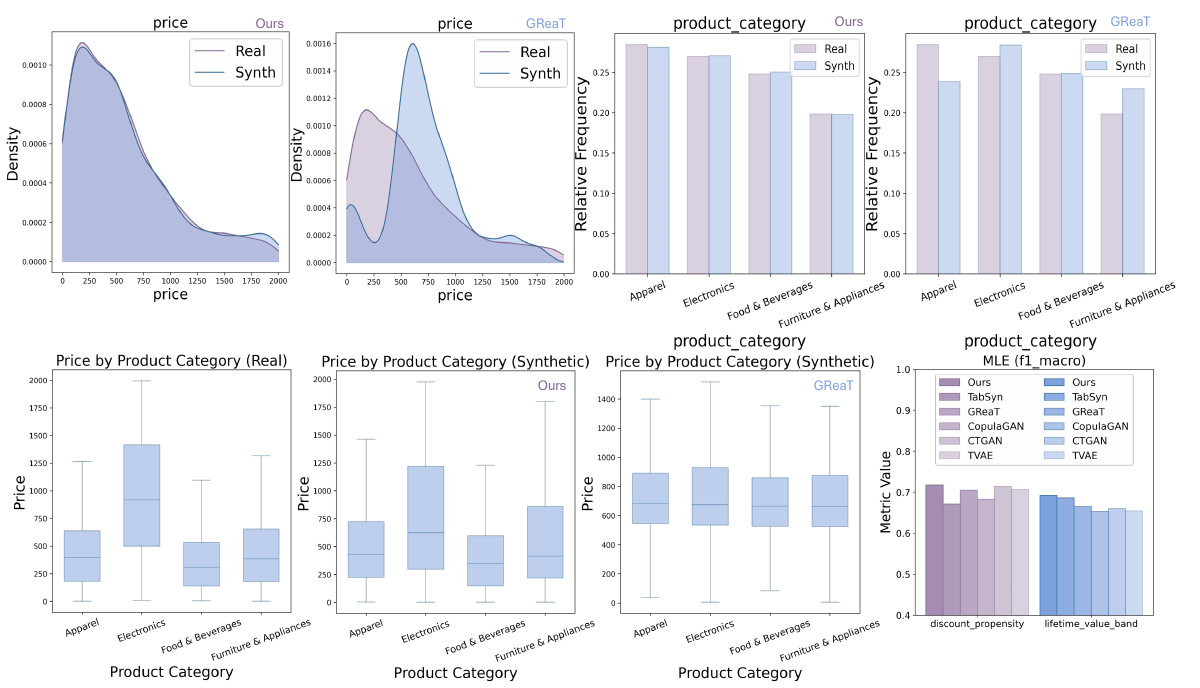}%
    \vspace{-10pt}
    \caption{Qualitative Distributions and Comparisons.}
    \label{fig:trans_main}
  \end{minipage}%
\vspace{-15pt}
\end{figure*}

\renewcommand{\arraystretch}{1.2}
\begin{table*}[h]
\resizebox{\linewidth}{!}{
\begin{tabular}{l|c|cccccc|ccc}
\hline
\multirow{2}{*}{\textbf{Methods}} & \multirow{2}{*}{\textbf{Rej. \%}} & \textbf{Age} & \textbf{Gender} & \textbf{Location} & \textbf{Category} & \textbf{Price} & \textbf{Payment} & \textbf{{[}v\_A, v\_G, v\_C{]}} & \textbf{{[}v\_C, v\_X{]}} & \textbf{{[}v\_L, v\_M{]}} \\
 &  & \textbf{Was} & \textbf{Tvd} & \textbf{Tvd} & \textbf{Tvd} & \textbf{Was} & \textbf{Tvd} & \textbf{Jsd} & \textbf{Jsd} & \textbf{Jsd} \\ \hline
\rowcolor{MyBlueA!25} \textbf{Ours (GPT-4.1-nano)} & \textbf{0.3 ± 0.1} & 1.13 & 0.002 & \textbf{0.002} & 0.022 & 12.762 & 0.003 & \textbf{0.071} & \textbf{0.134} & \textbf{0.007} \\
\rowcolor{MyBlueA!50} \textbf{Ours (Qwen2.5-7B)} & 0.5 ± 0.1 & 1.03 & \textbf{0.001} & 0.013 & 0.007 & \textbf{8.691} & \textbf{0.001} & 0.089 & 0.16 & 0.021 \\
\rowcolor{MyBlueA!100} \textbf{Ours (Qwen2.5-3B)} & 0.5 ± 0.1 & \textbf{0.93} & \textbf{0.001} & \textbf{0.002} & \textbf{0.003} & 14.367 & 0.002 & 0.082 & 0.198 & 0.139 \\ \hline
\end{tabular}
}
\vspace{3pt}
\caption{Quantitative ablation study on LLM backbones. We compare the synthesis quality of \model using GPT-4.1-nano, Qwen2.5-7B-Instruct, and Qwen2.5-3B-Instruct against the TabSyn baseline. The results highlight the framework's adaptability across models of varying parameter sizes, with bold values indicating the best performance in each category.}

\label{tab:backbone_abl}
\vspace{-20pt}

\end{table*}

Having demonstrated \model's unique capabilities, we now evaluate its statistical fidelity against strong, established baselines to answer \textmyblue{RQ2}. We consider a controlled e-commerce transaction synthesis task with a tractable likelihood, enabling a rigorous comparison with state-of-the-art \textmyblue{tabular} data synthesis models. This experiment is designed to test whether \model, operating \emph{solely} on macro-statistics, can match or exceed the performance of models trained with full access to micro-level records.

\noindent \textbf{Task Setup.}
We construct a controlled environment in which each transaction is generated from a known Bayesian network over six variables, $\{v_A, v_G, v_L, v_C, v_X, v_M\}$, corresponding to \texttt{user\_age}, \texttt{user\_gender}, \texttt{user\_location\_tier}, \texttt{product\_category}, \texttt{price}, and \texttt{payment\_method}. The underlying joint distribution follows a structured probabilistic graphical model: $p(v_A, v_G, v_L, v_C, v_X, v_M)
= p(v_A)\,p(v_G)\,p(v_L)\,p(v_C \mid v_A, v_G)\,p(v_X \mid v_C)\,p(v_M \mid v_L).$
This known generative process allows for a precise assessment of how well different synthesis methods capture ground-truth dependencies. We generate a reference dataset of 2{,}000 transactions from this model and treat it as ground truth. 
From this dataset, we derive the target macro-statistics including all marginal statistics and joint contingency tables over correlated variable groups defined by the Bayesian network.
These macro-statistics constitute the \emph{only} input provided to \model. In contrast, all baseline methods are trained directly on the full set of 2{,}000 micro-records. This design enables a direct and fair comparison of statistical fidelity, highlighting whether the joint dependencies inferred by \model from macro-statistics alone can recover the true underlying structure.

\noindent \textbf{Baselines.}
Given the structured tabular nature of the task, we compare \model against representative methods spanning major paradigms of tabular data synthesis, all of which are trained with full access to the micro-level records.
These include:
(1) VAE- and GAN-based models (TVAE, CTGAN);
(2) GANs with explicit dependency modeling (CopulaGAN);
(3) autoregressive transformers (GReaT);
(4) diffusion-based models (TabSyn);
(5) classical statistical alignment methods (IPF);
(6) hybrid LLM-probabilistic approaches (Spada, which combines LLM-induced dependency graphs with normalizing flows);
and (7) LLM-based in-context generation (LLM-ICL).
To ensure a fair comparison, we apply rejection sampling across all methods to enforce basic record-level validity.

\begin{figure}[h]  
  \centering
  \includegraphics[width=\columnwidth]{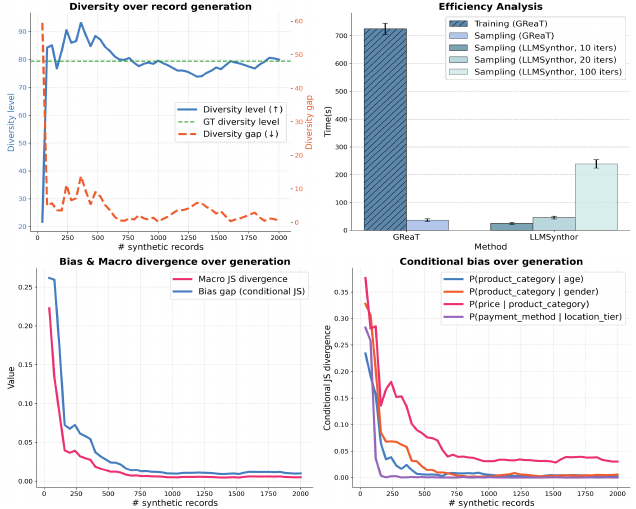}
  \vspace{-20pt}
  \caption{Diversity, macro-level alignment, and conditional bias rapidly converge toward ground truth as more synthetic records are generated, while \model also shows flexible efficiency, outperforming GReaT at low iteration counts and achieving the highest fidelity with more iterations.}
  \vspace{-8pt}
  \label{fig:diver}
\end{figure}
\noindent \textbf{Results.}
The results provide a clear affirmative answer to \textmyblue{RQ2}.
Despite operating solely on macro-statistics, \model consistently outperforms all baselines trained on the full micro-record dataset across both statistical fidelity and downstream utility.

We evaluate the synthesized data from two complementary perspectives.
\emph{First}, to assess statistical fidelity, Tables~\ref{tab:marginal_trans} and~\ref{tab:joint_trans} report performance on marginal and joint distributions using multiple metrics, including Wasserstein distance (W), Total Variation Distance (TVD), Jensen-Shannon Divergence (JSD), and the Classifier Two-Sample Test (C2ST) gap ($|\mathrm{acc}-0.5|$).
Across all metrics, \model achieves the lowest divergence scores, indicating the closest match to the ground-truth distributions.
The visualizations in Figure~\ref{fig:trans_main} further confirm that \model most faithfully preserves the target dependency structures.
\emph{Second}, we evaluate the practical downstream utility of the synthetic data.
We derive two rule-based downstream labels from the original transaction variables: \texttt{discount\_propensity}, which measures price sensitivity using category-normalized price deviation together with age, payment method, and location tier; and \texttt{lifetime\_value\_band}, which approximates customer value from transaction price, product category, payment method, and age. Both labels are computed using fixed, interpretable rules applied identically to real and synthetic data. We then train standard classifiers (logistic regression, decision trees, and random forests) on synthetic data generated by each method.
As shown in Figure~\ref{fig:trans_main}, models trained on \model's synthetic data generalize best to real data, demonstrating that its superior statistical fidelity translates directly into improved performance on downstream machine learning efficiency tasks.

\begin{table}[h]
\centering
\small
\renewcommand{\arraystretch}{.96}
\resizebox{\linewidth}{!}{
\begin{tabular}{lcccc}
\toprule
\textbf{Methods} & \textbf{R1 (\%) $\downarrow$} & \textbf{R2 (\%) $\downarrow$} & \textbf{R3 (\%) $\downarrow$} & \textbf{Rej. (\%) $\downarrow$} \\
\midrule
TVAE       & 1.6 $\pm$ 0.1 & 0.1 $\pm$ 0.1 & 0.1 $\pm$ 0.0 & 1.8 $\pm$ 0.1 \\
CTGAN      & 2.9 $\pm$ 0.1 & 1.9 $\pm$ 0.1 & 0.3 $\pm$ 0.0 & 5.1 $\pm$ 0.1 \\
CopulaGAN  & 2.1 $\pm$ 0.1 & 2.1 $\pm$ 0.1 & 0.6 $\pm$ 0.0 & 4.8 $\pm$ 0.1 \\
GReaT      & 3.5 $\pm$ 0.2 & 0.4 $\pm$ 0.0 & 0.0 $\pm$ 0.0 & 3.9 $\pm$ 0.1 \\
TabSyn     & 0.7 $\pm$ 0.1 & 0.1 $\pm$ 0.0 & 0.1 $\pm$ 0.0 & 0.8 $\pm$ 0.1 \\
Spada      & 3.6 $\pm$ 0.1 & 0.2 $\pm$ 0.0 & 0.4 $\pm$ 0.0 & 4.1 $\pm$ 0.1 \\
IPF        & 0.6 $\pm$ 0.3 & 0.5 $\pm$ 0.2 & 0.7 $\pm$ 0.0 & 1.8 $\pm$ 0.0 \\
LLM-ICL    & 0.0 $\pm$ 0.0 & 0.0 $\pm$ 0.0 & 0.0 $\pm$ 0.0 & 0.0 $\pm$ 0.0 \\ \midrule
\rowcolor{MyBlueA!50} GT (Real Data) & $0.5 \pm 0.1$ & $0.0 \pm 0.0$ & $0.1 \pm 0.0$ & $0.6 \pm 0.1$ \\
\rowcolor{MyBlueA!100}
\model     & \textbf{0.5 $\pm$ 0.1} & \textbf{0.0 $\pm$ 0.0} & \textbf{0.0 $\pm$ 0.0} & \textbf{0.5 $\pm$ 0.1} \\
\bottomrule
\end{tabular}
}
\vspace{2pt}
\caption{Rule-wise and total rejection rates (\%) across 5 runs of 2{,}000 generated e-commerce transactions per method.}
\vspace{-22pt}
\label{tab:ecom_rejection_rules}
\end{table}

We further evaluate semantic realism using three soft plausibility rules: \textbf{R1} rejects unrealistically low prices for large-item categories such as Electronics and Furniture \& Appliances; \textbf{R2} rejects extremely high Food \& Beverages spending by teenagers; and \textbf{R3} rejects ultra-expensive Electronics purchases by very senior users. These rules target rare but meaningful cross-variable inconsistencies in e-commerce transactions. As shown in Table~\ref{tab:ecom_rejection_rules}, \model achieves a $0.5\%$ rejection rate, matching the real-data reference ($0.6\%$) and outperforming most baselines. While LLM-ICL has a slightly lower rejection rate, its weaker distributional fit shows that semantic validity alone is insufficient, \model better balances plausibility and statistical fidelity.
Finally, as shown in Figure~\ref{fig:diver}, \model maintains high diversity while progressively mitigating bias. Efficiency analysis further demonstrates the framework's flexibility: with only 10 iterations, \model outperforms GReaT in both speed and quality. Increasing iterations further enhances fidelity, offering a controllable trade-off between computational cost and performance.  
In Table~\ref{tab:backbone_abl}, we conduct \textmyblue{robustness} checks demonstrating stable SOTA performance when using smaller, open-source models (Qwen2.5-Instruct with 3B and 7B).

\vspace{-6pt}
\subsection{Population Synthesis}

Having established \model's quantitative advantages on structured data, we now address \textmyblue{RQ3} by evaluating its ability to generalize to complex, unstructured data formats.
We consider a real-world population synthesis task, where records are inherently \textmyblue{unstructured} due to varying household sizes and hierarchical household-person relationships. This setting tests whether \model’s general-purpose, macro-aware framework can outperform specialized population synthesis models that typically rely on extensive task-specific modeling and manual constraints.

\noindent \textbf{Task Setup.} 
We use population microdata from the American Community Survey (ACS) for households in South Carolina. The dataset contains both household- and person-level attributes, resulting in variable-length records that pose a significant challenge for conventional tabular synthesis methods. After preprocessing, we obtain approximately 15{,}000 households. The objective is to generate a synthetic population that preserves complex joint distributions across demographic, socioeconomic, and household-structure variables. 
To evaluate practical fidelity, we define 16 policy-relevant queries (e.g., the proportion of multigenerational households) grouped into six thematic categories.
These queries serve as a proxy for real-world utility by evaluating whether the synthetic population preserves \emph{meaningful joint distributions} over combinations of demographic and household variables.

\noindent \textbf{Baselines.} 
We compare \model against strong, specialized population synthesis methods:
(1) CP, a tensor factorization approach;
(2) HMM, a specifically designed hierarchical probabilistic model for population synthesis; and
(3) NVI, a deep variational inference framework.
These baselines represent state-of-the-art techniques tailored specifically to population synthesis, providing a rigorous benchmark for evaluating the generality of \model.

\vspace{-6pt}
\begin{table}[h]
    \centering
    
    \renewcommand{\arraystretch}{1.0}
    \resizebox{\linewidth}{!}{
        \begin{tabular}{lccccccc}
        \toprule
        \textbf{Methods} & \textbf{Rej. Rate\%} & \textbf{Demog.} & \textbf{Employment} & \textbf{Equity} & \textbf{Household} & \textbf{Mobility} & \textbf{Vuln.} \\ \midrule
        CP  & 73.9 & 0.54 & 1.02 & 5.79 & 2.34 & 1.47 & 0.86 \\
        HMM & 57.8 & 0.56 & 0.32 & 4.23 & 2.01 & 0.48 & 0.91 \\
        NVI & 96.8 & 0.53 & 0.27 & 5.49 & 2.06 & \textbf{0.24} & 1.06 \\ \midrule
        \rowcolor{MyBlueA!100} Ours & \textbf{13.3} & \textbf{0.21} & \textbf{0.2} & \textbf{0.25} & \textbf{0.13} & 0.35 & \textbf{0.37} \\ \bottomrule
        \end{tabular}
    }
    \vspace{-2pt}
    \caption{Rejection rate and category-wise MRE across queries.}
    \label{tab:pop_utils}
    \vspace{-7pt}

    \renewcommand{\arraystretch}{1.05}
    \resizebox{\linewidth}{!}{
        \begin{tabular}{lccc>{\columncolor{MyBlueA!50}}c>{\columncolor{MyBlueA!100}}c}
        \toprule
        \textbf{Check} & \textbf{NVI} & \textbf{CP} & \textbf{HMM} & \textbf{Real} & \textbf{Ours} \\
        \midrule
        \multicolumn{1}{l}{\textbf{\textit{Overall rejection rates $\downarrow$}}} & & & & & \\
        \quad Household rejection rate & 96.8 $\pm$ 0.3 & 73.9 $\pm$ 0.6 & 57.8 $\pm$ 0.5 & 0.0 $\pm$ 0.0 & \textbf{13.3 $\pm$ 0.4} \\
        \quad Person rejection rate    & 45.9 $\pm$ 0.4 & 34.1 $\pm$ 0.7 & 27.8 $\pm$ 0.6 & 0.0 $\pm$ 0.0 & \textbf{6.3 $\pm$ 0.2}  \\
        \midrule
        \multicolumn{1}{l}{\textbf{\textit{Household-level checks $\downarrow$}}} & & & & & \\
        \quad householder\_adult\_consist. & 4.5 $\pm$ 0.2 & 4.4 $\pm$ 0.3 & 15.2 $\pm$ 0.5 & 0.0 $\pm$ 0.0 & \textbf{3.8 $\pm$ 0.2} \\
        \quad persons\_all\_valid                   & 96.7 $\pm$ 0.4 & 73.5 $\pm$ 0.6 & 56.1 $\pm$ 0.7 & 0.0 $\pm$ 0.0 & \textbf{10.7 $\pm$ 0.3} \\
        \midrule
        \multicolumn{1}{l}{\textbf{\textit{Person-level checks $\downarrow$}}} & & & & & \\
        \quad age\_range                            & 0.1 $\pm$ 0.0 & 0.5 $\pm$ 0.1 & 0.3 $\pm$ 0.1 & 0.0 $\pm$ 0.0 & \textbf{0.0 $\pm$ 0.0} \\
        \quad race\_valid                           & 0.0 $\pm$ 0.0 & 0.8 $\pm$ 0.1 & 0.0 $\pm$ 0.0 & 0.0 $\pm$ 0.0 & \textbf{0.0 $\pm$ 0.0} \\
        \quad employment\_age\_consist.          & 41.6 $\pm$ 0.7 & 30.9 $\pm$ 0.6 & 24.5 $\pm$ 0.5 & 0.0 $\pm$ 0.0 & \textbf{4.4 $\pm$ 0.2} \\
        \quad education\_age\_consist.           & 42.7 $\pm$ 0.8 & 8.1 $\pm$ 0.4  & 10.7 $\pm$ 0.5 & 0.0 $\pm$ 0.0 & \textbf{2.2 $\pm$ 0.1} \\
        \bottomrule
        \end{tabular}
    }
    \vspace{1pt}
    \caption{Rejection rates (\%) by validation rule across methods. Rules with zero rejections for all methods are omitted for brevity. ``Consist.'' denotes consistency checks.}
    \label{tab:rejection}
\vspace{-18pt}
\end{table}

\noindent \textbf{Results.} 
The results provide a clear affirmative answer to \textmyblue{RQ3}. As shown in Table~\ref{tab:pop_utils}, \model achieves the lowest mean relative error (MRE) across all six categories of policy-relevant queries. For example, on equity-related metrics, the MRE is reduced from 4.23 (HMM) to 0.25. Consistent improvements are also observed for demographic, employment, mobility, and vulnerability-related queries. While \model does not outperform all baselines on every individual query, its consistent category-level superiority indicates a stronger ability to capture the complex joint dependencies inherent in population data, leading to substantially higher practical utility. Beyond statistical accuracy, \model exhibits a significant advantage in micro-record realism. As shown in Table~\ref{tab:rejection}, population data is governed by strict logical constraints (e.g., a child cannot be designated as a householder), which many probabilistic models struggle to enforce. As a result, traditional baselines incur extremely high rejection rates, reaching 96.8\% for NVI and 73.9\% for CP. In contrast, \model reduces the overall rejection rate to 13.3\%. By leveraging the semantic reasoning capabilities encoded in LLMs, \model naturally respects complex intra-household rules, such as valid role assignments and age-consistent attributes, without requiring explicit constraint programming. 

\begin{figure}[h]
    \centering
    \vspace{-7pt}
    \includegraphics[width=\linewidth]{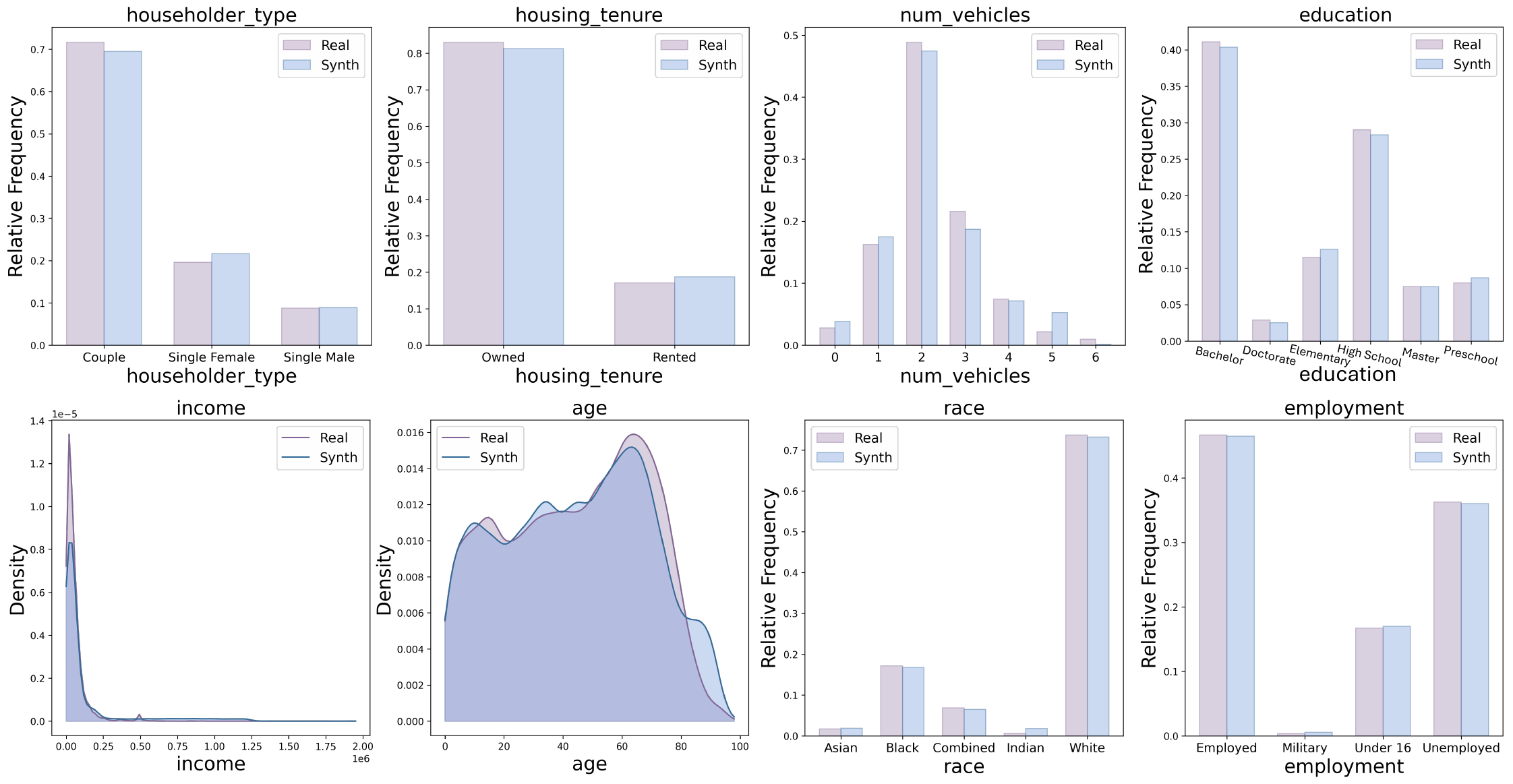}
    \vspace{-20pt}
    \caption{Generated population data by \model.}
    \label{fig:pop_ours_dist}
    \vspace{-9pt}
\end{figure}

Figure~\ref{fig:pop_ours_dist} compares the marginal distributions of real data and synthetic data generated by \model. \model closely matches the real distributions across both household-level and individual-level variables, including skewed variables such as \texttt{income} and \texttt{age}, as well as categorical variables such as \texttt{employment} and \texttt{race}. This suggests that the iterative structure-guided mechanism improves fidelity in challenging, high-variance settings.

\subsection{Discussion}

Overall, the experimental results provide strong evidence that \model effectively bridges the micro-macro gap in data synthesis.
Across diverse domains and data formats, \model can generate realistic micro-records using only multi-source macro-statistics (\textmyblue{RQ1}), achieve superior quantitative fidelity compared to baselines with full access to micro-level data (\textmyblue{RQ2}), and generalize to complex, unstructured population data without requiring task-specific engineering (\textmyblue{RQ3}).
Together, these results highlight the potential of macro-aware, LLM-driven synthesis as a general-purpose alternative to specialized data generation pipelines. Regarding privacy issues, \model operates exclusively on aggregate macro-statistics. By iteratively steering synthetic data toward population-level targets, the framework decouples generation from any individual record.
As a result, the synthesized data reflects distributional properties rather than specific individuals, substantially reducing the risk of direct re-identification.
This design aligns with ``privacy by aggregation'' principles and suggests a promising pathway for integrating formal privacy mechanisms, such as differential privacy applied at the macro-statistic level, in future extensions.

\section{Conclusion}

In this work, we introduced \model, a macro-aware framework that bridges the micro-macro gap in data synthesis. We address the fundamental challenge of generating realistic, individual-level micro-records when only aggregate macro-statistics are available. Our key contribution is a shift in how LLMs are used for data generation: rather than sampling records independently, we repurpose a pre-trained LLM as a macro-aware simulator that operates within a discrepancy-guided feedback loop, ensuring dataset-level statistical alignment.
Extensive experiments demonstrate that \model synthesizes realistic and useful micro-records from multi-source macro-statistics (\textmyblue{RQ1}), consistently outperforms state-of-the-art baselines trained on full micro-level data (\textmyblue{RQ2}), and generalizes effectively to complex, unstructured population data without requiring task-specific engineering (\textmyblue{RQ3}). Together, these results highlight the strength of macro-guided synthesis as a general and scalable alternative to specialized generative pipelines.
By enabling statistically grounded micro-record generation under severe data constraints, \model opens new opportunities for data-driven social science, agent-based simulation, and evidence-based policymaking. More broadly, this work suggests a principled pathway for combining the expressive priors of large language models with explicit statistical control, offering a foundation for building reliable, high-fidelity synthetic worlds across a wide range of scientific and societal applications.

\section{Limitations and Ethical Considerations}

Despite these strengths, several limitations remain. 
\textmyblue{First}, LLMs encode inherent semantic priors that may occasionally conflict with target macro-statistics, leading to biased or over-regularized generations; in practice, this effect can be mitigated through more constrained prompting or by selectively suppressing semantic cues. 
\textmyblue{Second}, the scalability of the framework is currently bounded by the context window and reasoning capacity of the underlying LLM, particularly in high-dimensional settings with many variables, though this limitation is expected to ease as LLM capabilities continue to advance. 
\textmyblue{Third}, while \model is well suited for mixed-type micro-records, it is not designed to directly model perceptual or tightly sequential data such as images, videos, graphs, or raw time series; nevertheless, it can naturally serve as a high-level macro-controller to guide domain-specific generators in these modalities. 
From an ethical and privacy perspective, \model operates exclusively on aggregate macro-statistics rather than individual-level records, decoupling generation from any specific individual and ensuring that synthesized data reflects distributional properties instead of memorizing or reproducing personal data.

\section*{GenAI Disclosure}
GPT-4.1-nano was used for output generation, and LLMs for language editing only; all study design, implementation, evaluation, findings, and interpretations were author-verified.

\section*{Acknowledgement}
This study was supported by the Natural Sciences and Engineering Research Council of Canada through NSERC ALLRP 602710-24.

\clearpage

\bibliographystyle{ACM-Reference-Format}
\bibliography{main}

\clearpage
\appendix            

\section{Implementation Details}
\label{appx:exp_ipl}

\paragraph{Macro-statistics}

We construct macro-statistics from the available target information. In real-world settings, these statistics can be directly specified by survey tables, census summaries, or other aggregate sources. In our experiments, because real micro-records are needed for baseline comparison, we derive the target macro-statistics from the ground-truth datasets. More generally, when micro-records are available, the same procedure can be used to obtain finer-grained macro-statistics for more sensitive discrepancy evaluation and correction.
For continuous variables, we use a hierarchical binning scheme. First, each variable is partitioned into a fixed number of quantile-based main bins, with 6 bins by default, so that each bin contains approximately the same number of real records. This reduces bin sparsity and gives stable coverage across the full distribution, including heavy tails and outliers. Second, to capture local mismatches, we identify the main bin with the largest positive discrepancy between real and synthetic data, i.e., where the real proportion most exceeds the synthetic proportion, and subdivide it into a fixed number of uniformly spaced sub-bins. The sub-bin frequencies are normalized to sum to the frequency of the parent bin.
This two-stage design captures both global and local distributional differences. Quantile-based main bins provide stable coarse statistics, while targeted sub-bin refinement focuses on regions where the synthetic data is most underrepresented. Discrete variables, as well as discretized continuous variables, are summarized as frequency tables. For joint macro-statistics, we compute empirical contingency tables over selected variable groups, using the same binning strategy for any continuous variables. 

\vspace{-7pt}
\paragraph{Discrepancy}

We measure the discrepancy $Q(\cdot,\cdot)$ between target and synthetic data as the entry-wise difference between their frequency tables, at both marginal and joint levels. For a target statistic $\phi$ and its synthetic estimate $\hat{\phi}$, we define Q$(\hat{\phi}, \phi) = \phi - \hat{\phi}.$
Since our synthesis process can only add records rather than remove them, we focus on positive discrepancies, corresponding to bins or contingency cells where the target proportion exceeds the synthetic proportion. These are the underrepresented regions that can be directly corrected through further generation.
Because the macro-statistics are built from interpretable bins, each discrepancy can be attributed to a specific value range for continuous variables, a category for discrete variables, or a combination of variables for joint statistics. This attributability is crucial for the iterative synthesis loop: it tells the LLM where the synthetic data underestimates the target distribution and guides it to generate corrective records in those regions, rather than making undirected global adjustments.

\vspace{-7pt}
\paragraph{Hyperparameters}

We generate 5 proposals per iteration for 100 iterations. At each iteration, 3 joint variable combinations are inferred for grounding. LLM inference is performed using GPT-4.1-nano with temperature 0.8, and Qwen2.5-7B-Instruct through the VLLM~\cite{kwon2023efficient} framework. The generation hyperparameters are: \texttt{max\_new\_tokens} = 2048, \texttt{temperature} = 0.7, \texttt{top\_k} = 20, and \texttt{top\_p} = 0.98.
For macro-statistics, when only partial target statistics are available, such as externally provided aggregate statistics, we compute discrepancies using those statistics directly. When full micro-record data is available, we apply the two-level quantile-based binning scheme described above, using 6 main bins and 8 sub-bins for continuous variables. Discrete variables and discretized continuous bins are handled uniformly when computing marginal and joint contingency tables. All experiments are repeated at least 3 times, and we report the mean results. The uncertainty in our results mainly comes from the stochasticity of LLM generation.

\vspace{-7pt}
\paragraph{Hardware}

All experiments are conducted on a server equipped with an Intel Xeon E5-2698 v4 CPU (40 threads), 252 GB of RAM, and four NVIDIA Tesla V100 GPUs with 32 GB of memory each.

\vspace{-3pt}
\section{Running Examples}
\label{appx:examples}

\subsection{Discrepancy-Guided Generation}

\label{sec:example_discrepancy}
This subsection explains how the model identifies the most significant discrepancies in each iteration and samples accordingly, thereby guiding the generation of micro-records to progressively reduce the gap between the synthetic and target distributions.

\begin{figure}[h]
  \centering
  \includegraphics[width=\linewidth]{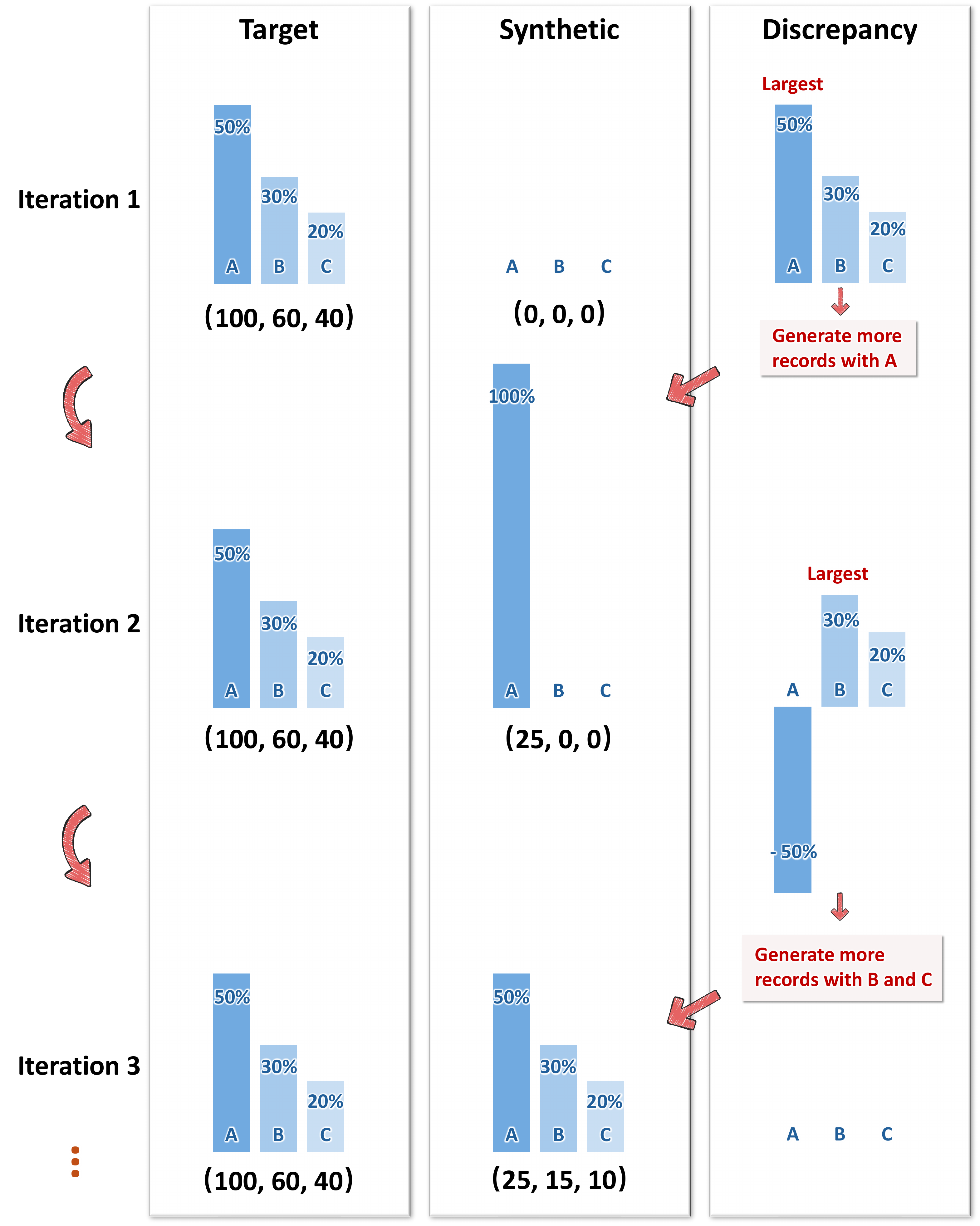}
  \vspace{-15pt}
  \caption{\textbf{Discrepancy-guided Iterative Synthesis.} The figure shows how discrepancies between target and synthetic distributions guide micro-record generation. At each iteration, the model samples from the largest positive discrepancies, highlighted in red, and generates records for underrepresented categories. Starting from an empty synthetic dataset \((0, 0, 0)\), the model first adds records for category A, then iteratively corrects categories B and C, progressively improving alignment with the target distribution.}
  \label{666}
\end{figure}

\subsection{Example of a Single Iteration}

\label{sec:example_iter}

We provide an example to show how each iteration improves the alignment between the synthetic and target distributions.

\begin{figure}[h]
  \centering
  \vspace{-5pt}
  \includegraphics[width=\linewidth]{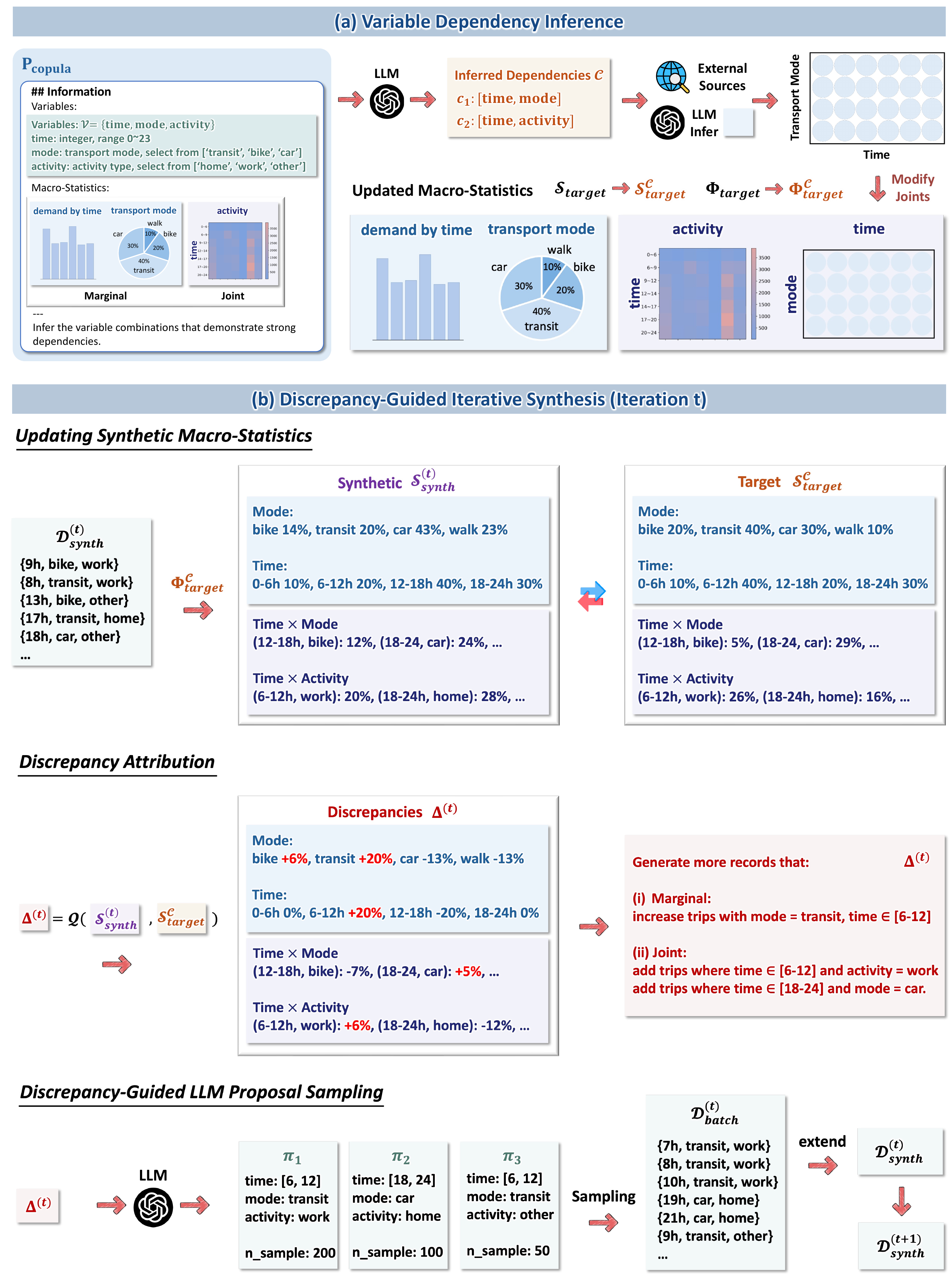}
  \vspace{-5pt}
  \caption{A simplified running example of discrepancy-guided iterative synthesis.}
  \label{fig:example_full}
\end{figure}

\section{Prompts}
\label{appx:prompts}

\subsection{LLM Proposal Sampling}
\label{sec:prompt_proposal}

\begin{tcolorbox}[title=$\texttt{p}_{\mathrm{proposal}}$: LLM Proposal Sampling, breakable]
\begin{minted}{markdown}
## Output Format:
```json
{{
  "n_proposals": n,
  "proposal1": {{
    "reason": "..." ,
    "proposal": "...",
    "num": n1
  }},
  "proposal2": {{ ... }}, ...
}}
```

## Information:
**Joint Guidance:** `{joint_guide}`
**Marginal Guidance:** `{marginal_guide}`
**Variables:** `{data_desc}`

---
## Rules:
Create less than {n_proposals} proposals totaling {n_samples} samples:
1. Each proposal must follow Joint and Marginal Guidance, do not improvise beyond the provided Guidance.
2. The reason for each proposal should explain the realistic meaning of this proposal and how it follows the provided Guidance by referencing frequencies one by one.
3. 'min', 'max', 'category' must use the actual values mentioned in **Variables**: Categorical variables must be a single valid candidate string (case-sensitive), and numerical variables must be a list of two integer or float numbers (e.g., [3.0, 5.1]).
4. If a variable value has a high frequency, that value should be selected in multiple proposals.
5. Most generated samples should, as much as possible, prioritize satisfying components that are common across the Guidances, and the “num” for each proposal should be determined based on the frequencies specified in the Guidance.

---  
Now return a pure, valid, and non-empty JSON in English that can be directly parsed by json.loads() in Python
\end{minted}
\end{tcolorbox}
\vspace{2pt}

In our implementation, each proposal defines a distribution for every variable. For discrete variables, it assigns a valid category, for continuous variables, it specifies a value range, such as [3.0, 5.1], from which values are sampled uniformly or by another simple scheme. The \texttt{num} field specifies how many records to generate from each proposal, allowing the LLM to allocate record counts according to the provided frequencies and guidance. In this way, the LLM can plan both diversity and statistical alignment by balancing the number and distribution of proposals. To make this process interpretable, we use chain-of-thought~\cite{wei2022chain} prompting, asking the LLM to explain the rationale for each proposal with reference to the given statistics and constraints.

This proposal format is one practical instantiation of LLM Proposal Sampling, but the framework is extensible. A proposal distribution can also be defined as executable code, tool calls, or pointers to external generators, such as ControlNet-based diffusion models for images or specialized LLM agents for domain-specific content. Thus, LLM Proposal Sampling can act as a high-level distributional controller that guides external generators or hybrid pipelines toward statistically faithful and scenario-aligned synthetic data across data types and domains.

\subsection{LLM as a Copula for Variable Dependency Inference}
\label{sec:prompt_copula}
\begin{tcolorbox}[title=$\texttt{p}_{\mathrm{copula}}$: LLM for Variable Dependency Inference, breakable]
\begin{minted}{markdown}
## Information
**Variables:** ```{data_desc}```

## Output Format:
{{
  "1": ["var1", "var2", ...], 
  "2": ["var1", "var2", ...], 
  ...
}}

### Example Output JSON:
{{
  "1": ["Temperature", "Humidity"], 
  "2": ["Humidity", "WeatherCondition"],
}}

---
Your task is to extract **at most {n_joints}** correlated variable groups based on the **Variables** summaries and present them in JSON format.
1. Ensure each group contains two or more variables. 
2. Format the correlated variable groups according to the **Output Format**.
\end{minted}
\end{tcolorbox}

\end{document}